\documentclass[sigconf, nonacm]{acmart}

\usepackage{multirow}
\usepackage{subfigure}
\usepackage{algorithm}
\usepackage[noend]{algorithmic}
\usepackage{xspace}
\usepackage{dsfont}
\usepackage{epsfig}
\usepackage{epstopdf}
\usepackage{booktabs}
\usepackage{indentfirst}
\usepackage{amsmath}
\usepackage{amsthm}
\usepackage{bm}
\usepackage{listings}
\usepackage{tabu}
\usepackage{longtable}
\usepackage{threeparttable}
\usepackage{marvosym}
\usepackage{color}
\usepackage{tabularx}
\usepackage{hyperref}
\usepackage{xspace}
\usepackage{multirow}
\usepackage{subfigure}
\usepackage{algorithm}
\usepackage{xspace}
\usepackage{dsfont}
\usepackage{epsfig}
\usepackage{epstopdf}
\usepackage{booktabs}
\usepackage{indentfirst}
\usepackage{amsmath}
\usepackage{amsthm}
\usepackage{amsfonts}
\usepackage{bm}
\usepackage{listings}
\usepackage{tabu}
\usepackage{longtable}
\usepackage{threeparttable}
\usepackage{marvosym}
\usepackage{tabularx}
\usepackage{color}
\usepackage{enumerate}   
\usepackage{enumitem}
\usepackage{tcolorbox}
\usepackage{balance}

\newcommand{\ie}{\emph{i.e.,}\xspace}

\newcommand{\eg}{\emph{e.g.,}\xspace}

\newcommand{\ignore}[1]{}

\newcommand\vldbdoi{XX.XX/XXX.XX}
\newcommand\vldbpages{XXX-XXX}
\newcommand\vldbvolume{14}
\newcommand\vldbissue{1}
\newcommand\vldbyear{2020}
\newcommand\vldbauthors{\authors}
\newcommand\vldbtitle{\shorttitle} 
\newcommand\vldbavailabilityurl{https://github.com/LibCity}
\newcommand\vldbpagestyle{plain} 

\begin{document}
\title{Unified Data Management and Comprehensive Performance Evaluation for Urban Spatial-Temporal Prediction [Experiment, Analysis \& Benchmark]}

\author{Jiawei Jiang}
\email{jwjiang@buaa.edu.cn}
\affiliation{%
  \institution{School of Computer Science and Engineering, Beihang University}
  \city{Beijing}
  \country{China}
}

\author{Chengkai Han}
\email{ckhan@buaa.edu.cn}
\affiliation{%
  \institution{School of Computer Science and Engineering, Beihang University}
  \city{Beijing}
  \country{China}
}

% \author{Wenjun Jiang}
% \email{jwjun@buaa.edu.cn}
% \affiliation{%
%   \institution{School of Computer Science and Engineering, Beihang University}
%   \city{Beijing}
%   \country{China}
% }

\author{Wayne Xin Zhao}
\email{batmanfly@gmail.com}
\affiliation{%
  \institution{Gaoling School of Artificial Intelligence, Renmin University of China}
  \city{Beijing}
  \country{China}
}

\author{Jingyuan Wang}
\authornote{Corresponding author}
\email{jywang@buaa.edu.cn}
\affiliation{
  \institution{School of Computer Science and Engineering, School of Economics and Management, Beihang University}
  \city{Beijing}
  \country{China}
}
% \affiliation{
%   \institution{School of Economics and Management, Beihang University}
%   \city{Beijing}
%   \country{China}
% }

% \affiliation{%
%   \state{Ireland}

% \orcid{0000-0002-1825-0097}
% \affiliation{%
%   \country{Iceland}

% \orcid{0000-0001-5109-3700}
% \affiliation{%
% \email{vb@rocquencourt.com}

% \author{J\"org von \"Arbach}
% \affiliation{%
%   \institution{University of T\"ubingen}
%   \city{T\"ubingen}
%   \country{Germany}
% }
% \email{jaerbach@uni-tuebingen.edu}
% \email{myprivate@email.com}
% \email{second@affiliation.mail}

% \author{Wang Xiu Ying}
% \author{Zhe Zuo}
% \affiliation{%
%   \institution{East China Normal University}
%   \city{Shanghai}
%   \country{China}
% }
% \email{firstname.lastname@ecnu.edu.cn}

% \author{Donald Fauntleroy Duck}
% \affiliation{%
%   \institution{Scientific Writing Academy}
%   \city{Duckburg}
%   \country{Calisota}
% }
% \affiliation{%
%   \institution{Donald's Second Affiliation}
%   \city{City}
%   \country{country}
% }
% \email{donald@swa.edu}

\begin{abstract}
The field of urban spatial-temporal prediction is advancing rapidly with the development of deep learning techniques and the availability of large-scale datasets. However, challenges persist in accessing and utilizing diverse urban spatial-temporal datasets from different sources and stored in different formats, as well as determining effective model structures and components with the proliferation of deep learning models. This work addresses these challenges and provides three significant contributions. Firstly, we introduce "atomic files", a unified storage format designed for urban spatial-temporal big data, and validate its effectiveness on 40 diverse datasets, simplifying data management. Secondly, we present a comprehensive overview of technological advances in urban spatial-temporal prediction models, guiding the development of robust models. Thirdly, we conduct extensive experiments using diverse models and datasets, establishing a performance leaderboard and identifying promising research directions. Overall, this work effectively manages urban spatial-temporal data, guides future efforts, and facilitates the development of accurate and efficient urban spatial-temporal prediction models. It can potentially make long-term contributions to urban spatial-temporal data management and prediction, ultimately leading to improved urban living standards.
\end{abstract}

% As deep learning technology advances and more urban spatial-temporal data accumulates, an increasing number of deep learning models are being proposed to solve urban spatial-temporal prediction problems. However, there are limitations in the existing field, including open-source data being in various formats and difficult to use, few papers making their code and data openly available, and open-source models often using different frameworks and platforms, making comparisons challenging. A standardized framework is urgently needed to implement and evaluate these methods. To address these issues, we provide a comprehensive review of urban spatial-temporal prediction and propose a unified storage format for spatial-temporal data called \textit{atomic files}. We also propose \name, an open-source library that offers researchers a credible experimental tool and a convenient development framework. In this library, we have reproduced \modelCnt spatial-temporal prediction models and collected \rawDataCnt spatial-temporal datasets, allowing researchers to conduct comprehensive experiments conveniently. Using \name, we conducted a series of experiments to validate the effectiveness of different models and components, and we summarized promising future technology developments and research directions for spatial-temporal prediction. By enabling fair model comparisons, designing a unified data storage format, and simplifying the process of developing new models, \name is poised to make significant contributions to the spatial-temporal prediction field.

\maketitle

\pagestyle{\vldbpagestyle}
\begingroup\small\noindent\raggedright\textbf{PVLDB Reference Format:}\\
\vldbauthors. \vldbtitle. PVLDB, \vldbvolume(\vldbissue): \vldbpages, \vldbyear.\\
\href{https://doi.org/\vldbdoi}{doi:\vldbdoi}
\endgroup
\begingroup
\renewcommand\thefootnote{}\footnote{\noindent
This work is licensed under the Creative Commons BY-NC-ND 4.0 International License. Visit \url{https://creativecommons.org/licenses/by-nc-nd/4.0/} to view a copy of this license. For any use beyond those covered by this license, obtain permission by emailing \href{mailto:info@vldb.org}{info@vldb.org}. Copyright is held by the owner/author(s). Publication rights licensed to the VLDB Endowment. \\
\raggedright Proceedings of the VLDB Endowment, Vol. \vldbvolume, No. \vldbissue\ %
ISSN 2150-8097. \\
\href{https://doi.org/\vldbdoi}{doi:\vldbdoi} \\
}\addtocounter{footnote}{-1}\endgroup

\ifdefempty{\vldbavailabilityurl}{}{
\vspace{.3cm}
\begingroup\small\noindent\raggedright\textbf{PVLDB Artifact Availability:}\\
The source code, data, and/or other artifacts have been made available at \url{\vldbavailabilityurl}.
\endgroup
}

\section{Introduction}\label{intro}
% \balance

In recent years, the advancements in sensor technology in urban areas have facilitated the collection of a vast amount of urban spatial-temporal data. This data offers new perspectives for leveraging artificial intelligence technologies to address urban spatial-temporal prediction challenges~\cite{yin2021deep}. Urban spatial-temporal prediction plays a crucial role in urban computing, enabling efficient management and decision-making in smart cities. Various applications benefit from urban spatial-temporal prediction, including congestion control~\cite{control}, route planning~\cite{routeplanning}, vehicle dispatching~\cite{dispatching}, and POI recommendation~\cite{POIrec}.

The fundamental distinction between urban spatial-temporal data and conventional time-series data lies in that spatial-temporal data comprises historical state time series from multiple spatial entities that exhibit mutual influences. Consequently, the spatial-temporal prediction problem can be formulated as a multivariate time series prediction problem~\cite{METRO, AutoCTS} rather than a univariate time series prediction problem~\cite{OneShotSTL}. Effectively modeling the spatial relationships among the time series of different entities stands as the key to solving the spatial-temporal prediction problem.

Traditional time series forecasting methods, such as VAR~\cite{VAR} and ARIMA~\cite{ARIMA}, fail to consider the spatial dependencies inherent in urban spatial-temporal data. In contrast, deep learning methods exhibit superior feature learning capabilities and are more effective in capturing spatial-temporal correlations. Consequently, numerous deep learning-based urban traffic prediction techniques have been proposed in the literature. Initially, researchers utilized convolutional neural networks (CNNs) to model spatial dependencies within Euclidean neighborhoods. This approach represented cities as grids, with CNNs employed to capture relationships among the time series of grid cells. As research advanced, the focus shifted towards modeling relationships between time series observed across different variables using a graph structure. Graph convolutional networks (GCNs) emerged as a popular choice for capturing spatial dependence in urban spatial-temporal data. Learning graph structures~\cite{wang2017community} has become a mainstream approach in spatial-temporal prediction, and spatial attention mechanisms have also been explored to capture the dynamic graph structures among variables.

The field of urban spatial-temporal prediction is experiencing rapid advancements driven by the development of deep learning techniques and the availability of large-scale datasets. However, several challenges still need to be addressed: Firstly, accessing and utilizing existing urban spatial-temporal datasets can be challenging due to their diverse sources and formats. These datasets are often stored in different formats, such as NPZ, PKL, H5, CSV, and soon. This heterogeneity creates barriers for users to effectively utilize and explore the datasets, particularly for newcomers in the field. This hinders the development and standardization of the field and presents obstacles to efficient data management and utilization.

Secondly, with the proliferation of deep learning models in the domain, it becomes increasingly difficult to determine effective model structures and designs. Deep learning models have demonstrated state-of-the-art performance in spatial-temporal prediction, leading to a surge in research papers in this area. However, as models become more complex and diverse, it is challenging to identify which techniques are truly effective and which directions hold potential for future research. The sensitivity of deep learning models to experimental parameters and configurations further emphasizes the need to explore valuable research directions and identify the most promising approaches. To address the challenges above, we make the following contributions in this paper:

% To address these challenges, We first define a unified storage format for urban spatial-temporal big data, called atomic files. Then, we provide a review of the technological advances in the field of urban spatial-temporal prediction, providing a comprehensive summary of the development of deep learning techniques within the field. Finally, we conduct comprehensive experiments to explore the effects of different models and components with the help of the \name framework~\cite{libcity}, an open-source library proposed in our past work to address the problem of reproducibility in the field of urban spatial-temporal prediction. In \name, we reproduce all models using Pytorch~\cite{pytorch}, allowing for a fair comparison of models. 

(1) \textbf{Unified Storage Format}: We introduce a unified storage format called "atomic files" specifically designed for urban spatial-temporal big data. Through the validation of 40 diverse datasets, we demonstrate the effectiveness of this format in simplifying the management of urban spatial-temporal data. This contribution addresses the challenge of accessing and utilizing diverse datasets stored in different formats, enables accurate and efficient urban prediction, and enhances related applications.

(2) \textbf{Technical Development Roadmap}: We present a comprehensive overview of the technological advances in urban spatial-temporal prediction models. This roadmap outlines techniques for effectively modeling spatial dependencies, temporal dependencies, and spatial-temporal fusion using deep learning. By providing this roadmap, we facilitate the development of advanced and robust urban spatial-temporal prediction models, empowering researchers to fully leverage the potential of deep learning in this domain.

(3) \textbf{Extensive Experiments and Performance Evaluation}: We conduct extensive experiments using 18 models and 20 datasets, establishing a performance leaderboard and comparing previous works in spatial-temporal prediction. Through these experiments, we gain valuable insights into model performance and identify promising directions for future research. This evaluation enables researchers and practitioners to make informed decisions and drive advancements in urban spatial-temporal prediction.

By introducing the atomic files storage format, providing a technical roadmap, and conducting comprehensive experiments, we contribute to effectively managing urban spatial-temporal data, guiding future research efforts, and facilitating the development of accurate and efficient urban spatial-temporal prediction models. These contributions aim to address the challenges faced by the field and accelerate progress in urban spatial-temporal prediction, ultimately improving the quality of urban living.

The subsequent sections are organized as follows: Section~\ref{data} introduces the unified storage format for spatial-temporal data. Section~\ref{tasks} presents the problem formalization of urban spatial-temporal prediction. Section~\ref{model} outlines the development roadmap for deep models in urban spatial-temporal prediction. Section~\ref{experiments} describes the comprehensive experiments conducted to compare different models and gain insights. Section~\ref{conclusion} provides the conclusion.

\section{Unified Spatial-Temporal Data Management} \label{data}
% \balance

With the rapid progress of technology, a large amount of urban spatial-temporal data has been accumulated, offering fresh insights for addressing urban traffic prediction challenges. Common urban spatial-temporal data include traffic state data, trajectory data, meteorological data, air quality data, etc. Nevertheless, these spatial-temporal data often exist in various formats, posing additional obstacles for researchers aiming to harness the full potential of such data. In this section, we present a unified storage format designed specifically for urban spatial-temporal big data and demonstrate its efficacy through validation using 40 diverse datasets. We firmly believe that this format will significantly simplify the management of urban spatial-temporal data.

\subsection{Spatial-Temporal Data Unit} \label{data_unit}
In this section, we analyze the basic units underlying the urban spatial-temporal data.

\begin{figure}[t]
    \centering
    \subfigure[Irregular Regions in NYC]{
        \includegraphics[width=0.45\columnwidth, page=1]{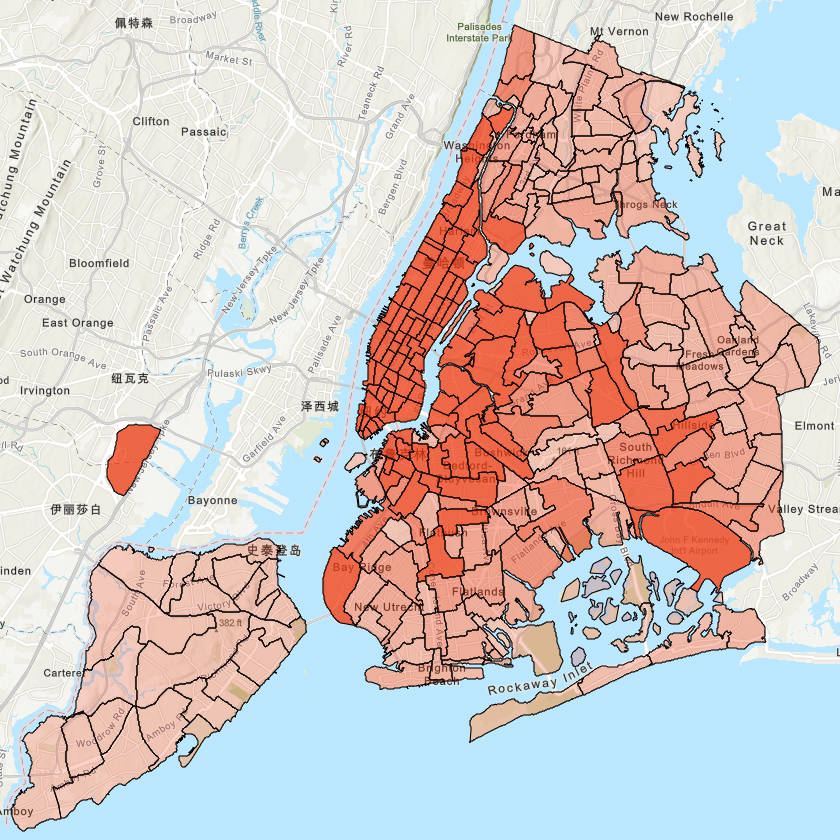}
        \label{fig:introa}
    }
    \subfigure[Regular Grids in Beijing]{
        \includegraphics[width=0.45\columnwidth, page=2, page=1]{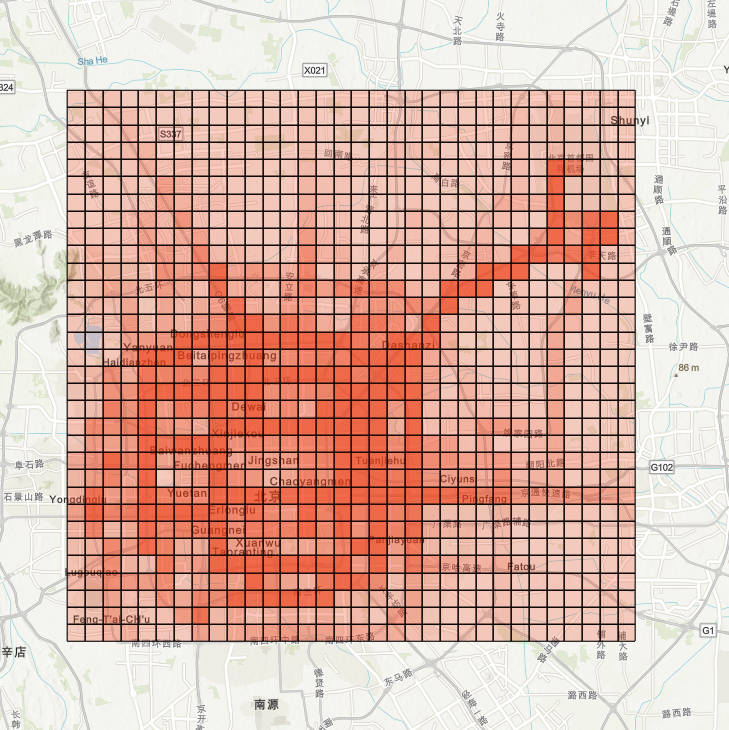}
        \label{fig:introb}
    }
    \vspace{-.4cm}
    \caption{Different division methods of plane-based data.}
    \label{fig:plane}
    \vspace{-.6cm}
\end{figure}

\subsubsection{Basic Unit}
The most basic unit in spatial-temporal data is the geographical unit. According to the difference in spatial distribution, the basic geographical unit data are divided into three types:

(1) \textbf{Point-based Data}: This type of data mainly includes urban point of interest (POI) data, GPS sampling points, urban traffic sensors, traffic cameras, and other device data. We can use $(geo_i, \boldsymbol{l}_i, \boldsymbol{p}_i)$ to represent the point data, where $geo_i$ is the ID of this point, $\boldsymbol{l}_i$ is the location of the point $i$, such as latitude and longitude information, and $\boldsymbol{p}_i$ is the attribute information of the point $i$, such as the POI category.

(2) \textbf{Line-based Data}: This type of data is mainly the road segment data. Similarly, we can use $(geo_i, \boldsymbol{l}_i, \boldsymbol{p}_i)$ to represent the line-based data, where $geo_i$ is the ID of the road segment, $\boldsymbol{l}_i$ is the location of the road segment, which usually includes the latitude and longitude of the starting and ending locations, and $\boldsymbol{p}_i$ is the attributes of road segment $i$, such as the average speed.

(3) \textbf{Plane-based Data}: This type of data is mainly urban region data such as administrative districts. In particular, the city grid data obtained by the grid division of the city area is also plane-based data. The different methods of dividing the city area are given in Figure~\ref{fig:plane}. Similarly, we can use $(geo_i, \boldsymbol{l}_i, \boldsymbol{p}_i)$ to represent the plane-based data, where $geo_i$ is the ID of the region, $\boldsymbol{l}_i$ is the location of this region, which usually includes the latitude and longitude of the regional boundary locations, and $\boldsymbol{p}_i$ is the attribute information of region $i$, such as the traffic inflow or outflow of the region.

\subsubsection{Unit Relations}
The unit relation data describes the relationships between geographical units in urban space, \eg road connectivity and distance relationships. Given that these relationships exhibit diverse structures, we further refine the interconnections between geographical units in the following manner:

% \begin{itemize}
%     \item Relations between Geographical Units: They describe the physical relations between geographical units in urban space, \eg road connectivity and distance relationships between geographical units.
%     \item Relations between User Units: They describe the social relationship between user units, \eg the friend relationship between user units.
%     \item Relations between Geographical Units and User Units: They describe the attribution relationship between geographical units and user units, \eg the relationship between the user's residence and workplace.
% \end{itemize}

% In particular, since the relationships between geographic units have multiple structures, we further refine the relationships between geographic units as follows:

(1) \textbf{Graph Relation Data}: Graph network data includes urban sensor graph, POI graph, road network graph, region graph, and soon. In general, we can define such graph data as $\mathcal{G} = (\mathcal{V}, \mathcal{E}, \boldsymbol{F}_\mathcal{V}, \boldsymbol{A})$, where $\mathcal{V}$ is a set of $N$ nodes ($N = |\mathcal{V}|$), $\mathcal{E} \subseteq \mathcal{V} \times \mathcal{V}$ is a set of edges, $\boldsymbol{F}_\mathcal{V} \in \mathbb{R}^{N \times D}$ is a $D$-dimension attribute matrix, and $\boldsymbol{A} \in \mathbb{R}^{N \times N}$ is a weighted adjacency matrix between nodes. % $\mathcal{N}_i$ is the neighborhood of the node $v_i$, 

(2) \textbf{Grid Relation Data}: A common Euclidean neighborhood data is the city grid data, which can be represented as $\boldsymbol{F}_{\boldsymbol{V}} \in \mathbb{R}^{I \times J \times D}$, where the city is divided into ${I \times J}$ grids, and each grid ${V}^{(i, j)}$ has $D$ attributes. Each grid matches the distribution of the surrounding grids in the Euclidean neighborhood.

(3) \textbf{Graph OD Relation Data}: This kind of data focuses on the relation between the origin and the destination, which is a more fine-grained relational data. We use $\boldsymbol{F}_\mathcal{E} \in \mathbb{R}^{N \times N \times D}$ to denote the $D$-dimension OD data between each pair of nodes in $\mathcal{V}$ (or say the $D$-dimension data of each edge), where each entry $\boldsymbol{F}_\mathcal{E}^{(i, j)} \in \mathbb{R}^D$ represents $D$-dimension attributes from node $v_i$ to node $v_j$. 

(4) \textbf{Grid OD Relation Data}: In particular, for the OD data between city grids, it can be denoted by $\boldsymbol{F}_\mathcal{E} \in \mathbb{R}^{I \times J \times I \times J \times D}$, and each entry $\boldsymbol{F}_\mathcal{E}^{(i_1, j_1), (i_2, j_2)}$ represents $D$-dimension attributes from $Grid^{(i_1, j_1)}$ to $Grid^{(i_2, j_2)}$.

Let us clarify the difference between Graph Relation Data and Graph OD Relation Data. Graph Relation Data pertains to the attributes of the graph nodes, \ie $\boldsymbol{F}_\mathcal{V} \in \mathbb{R}^{N \times D}$, such as the traffic speed of each node, in a graph network $\mathcal{G}$. On the other hand, Graph OD Relation Data is concerned with the attributes of the graph edges, \ie $\boldsymbol{F}_\mathcal{E} \in \mathbb{R}^{N \times N \times D}$, which represents the transfer flow from the start to the end of the edge. This is typically used to model traffic flow between origin-destination pairs in transportation networks. Similarly, for Grid Relation Data, we can refine it into OD data between each pair of grids, where the attributes of the edges represent the traffic flow between the grids.

\subsubsection{Unit Dynamics} \label{stdynamicdata}
Spatial-temporal dynamic data refers to the attribute information of city geographical units that change over time. We can obtain complex spatial-temporal data by incorporating dynamic attributes in the temporal dimension into the unit relation data. Examples of such data include urban traffic flow, traffic speed, and ridership demand, all exhibiting dynamic spatial-temporal characteristics. To illustrate this, let's consider graph relation data. When the graph attribute $\boldsymbol{F}_\mathcal{V} \in \mathbb{R}^{N \times D}$ changes over time, we can represent the $D$-dimensional spatial-temporal attributes of $N$ nodes at $T$ time slices using a spatial-temporal tensor denoted as $\boldsymbol{X} \in \mathbb{R}^{T \times N \times D}$. To comprehensively understand different data types, please refer to Table~\ref{tab:stdata} to explore the intuitive relationship between unit relations and unit dynamics.

\begin{table}[t]
% \small
  \centering
  \caption{Symbols for Unit Relations and Unit Dynamics}
  \resizebox{\columnwidth}{!}{
    \begin{tabular}{c|c|c}
    \toprule
    \textbf{Types}     &   \textbf{Unit Relations}    & \textbf{Unit Dynamics} \\
    \midrule
    Graph Relation Data     &   $\boldsymbol{F}_\mathcal{V} \in \mathbb{R}^{N \times D}$    & $\boldsymbol{X} \in \mathbb{R}^{T \times N \times D}$ \\
    \midrule
    Grid Relation Data     &   $\boldsymbol{F}_{\boldsymbol{V}} \in \mathbb{R}^{I \times J \times D}$    & $\boldsymbol{X} \in \mathbb{R}^{T \times I \times J \times D}$ \\
    \midrule
    Graph OD Relation Data     &   $\boldsymbol{F}_\mathcal{E} \in \mathbb{R}^{N \times N \times D}$    & $\boldsymbol{X} \in \mathbb{R}^{T \times N \times N \times D}$ \\
    \midrule
    Grid OD Relation Data     &   $\boldsymbol{F}_\mathcal{E} \in \mathbb{R}^{I \times J \times I \times J \times D}$    & $\boldsymbol{X} \in \mathbb{R}^{T \times I \times J \times I \times J \times D}$ \\
    \bottomrule
    \end{tabular}%
    }
    \vspace{-.5cm}
  \label{tab:stdata}%
  \vspace{-.2cm}
\end{table}%

\subsubsection{External Information}
External data refers to environmental information that provides additional context related to spatial-temporal dynamics. These data sources offer auxiliary information that can enhance the accuracy of urban spatial-temporal predictions. Examples of external data include city traffic accidents, major events, and calendar data, which encompass factors such as the day of the week, time of day, and the presence of holidays. By incorporating these external sources, we can leverage their valuable insights to improve the precision of spatial-temporal predictions. % in urban contexts.

% Table generated by Excel2LaTeX from sheet 'Sheet3'
% \begin{table*}[t]
%   \centering
%   \caption{Summary of Atomic Files}
%   \resizebox{\textwidth}{!}{
%     \begin{tabular}{c|c|c}
%     \toprule
%     \textbf{Suffix} & \textbf{Content} & \textbf{Format} \\
%     \midrule
%     .geo  & Geographical Unit Data & geo\_id, type, coordinates, properties \\
%     .usr  & User Unit Data & usr\_id, properties \\
%     .rel  & Unit Relation Data & rel\_id, type, origin\_id, des\_id, properties \\
%     .dyna & SDTD Data for User Unit (Trajectories) & dyna\_id, type, time, entity\_id, properties \\
%     .dyna & SSTD Data of Graph Network Relation Data & dyna\_id, type, time, entity\_id, properties \\
%     .grid & SDTD Data for Grid Relation Data & dyna\_id, type, time, row\_id, col\_id, properties \\
%     .od   & SDTD Data for OD Relation Data & dyna\_id, type, time, origin\_id, des\_id, properties \\
%     .gridod & SDTD Data for Grid OD Relation Data & dyna\_id, type, time, origin\_row\_id, origin\_col\_id, des\_row\_id, des\_col\_id, properties \\
%     .ext  & External Information & ext\_id, time, properties \\
%     \bottomrule
%     \end{tabular}
%     }
%   \label{tab:atomic_files}%
% \end{table*}%

\subsection{Atomic files} \label{atomic_files}
Existing open-source spatial-temporal datasets are often stored in various formats. This situation inevitably creates challenges and burdens for users who wish to utilize these datasets. To address this issue, based on the analysis of spatial-temporal data units, relations, and dynamics discussed in the previous section,  we propose the concept of "atomic files" as a solution, which offers a unified storage and management format for urban spatial-temporal data. These atomic files represent the minimum information units of urban spatial-temporal data. They can be classified into four types: Geographical Unit Data, Unit Relation Data, Spatial-temporal Dynamic Data, and External Data (Optional).

\begin{table*}[htbp]
  \centering
  \caption{Summary of Atomic Files}
    \begin{tabular}{c|c|c}
    \toprule
    \textbf{Suffix} & \textbf{Content} & \textbf{Format} \\
    \midrule
    .geo  & Geographical Unit Data & geo\_id, type, coordinates, [properties] \\
    .rel  & Unit Relation Data & rel\_id, origin\_id, des\_id, [properties] \\
    .dyna & Graph Relation Dynamic Data & dyna\_id,  time, \underline{entity\_id}, [properties] \\
    .grid & Grid Relation Dynamic Data & dyna\_id, time, \underline{row\_id, col\_id}, [properties] \\
    .od   & Graph OD Relation Dynamic Data & dyna\_id, time, \underline{origin\_id, des\_id}, [properties] \\
    .gridod & Grid OD Relation Dynamic Data & dyna\_id, time, \underline{origin\_row\_id, origin\_col\_id, des\_row\_id, des\_col\_id}, [properties] \\
    .ext  & External Information & ext\_id, time, [properties] \\
    \bottomrule
    \end{tabular}%
    \vspace{-0.5cm}
  \label{tab:atomic_files}%
  % \vspace{-0.1cm}
\end{table*}%

We adopt a comma-separated value format, similar to the CSV file, for storing the four types of atomic files. Each atomic file comprises multiple columns of data. To differentiate between different types of atomic files, we assign specific file suffixes. Additionally, we enforce specific guidelines for the information contained in each line of the atomic files, as outlined in Table~\ref{tab:atomic_files}. The precise definitions are as follows:

(1) The ".geo" tables store the Geographical Unit Data, which are required to include geo\_id (the primary key), geographic unit type (point, line, or polygon), and coordinate information (such as longitude and latitude). Other additional properties should be stored after the three columns above, such as the category of the point of interests (POIs) or the length of the road segments.

(2) The ".rel" tables store the Unit Relation Data, which should consist of rel\_id (the primary key), origin\_id, and des\_id. The origin\_id and des\_id represent the IDs of the origin and destination entities in the relationship, respectively. These IDs are derived from the geo\_id column (the foreign key) in the ".geo" table. Optionally, other properties can be included to indicate the weight of the relationship. Each row in the ".rel" table represents a directed edge from the origin to the destination.

(3) The Spatial-temporal Dynamic Data is stored in the ".dyna", ".grid", ".od", and ".gridod" tables, which are further divided into four sub-files due to the different unit relations. Each row in these tables represents an attribute of a geographic entity (or entity relationship) collected at a specific time. For instance, the dyna table stores spatial-temporal dynamic data related to graph relation data. It is required to include dyna\_id (the primary key), time, entity\_id, and at least one attribute, such as air quality and traffic flow. The time column represents the timestamp of the data collection, and the entity\_id column indicates the source of data collection. The main distinction among these four types of tables lies in the entity\_id column. In the dyna table, the entity\_id is derived from the geo\_id. In the grid table, the entity\_id is derived from a grid's row and column ID corresponding to the geo\_id  for simplicity. In the ".od" table, the entity\_id represents the origin\_id and des\_id of the OD relation entity. Lastly, in the ".gridod" table, the entity\_id corresponds to the row and column ID of a grid OD relation entity.

(4) The ".ext" tables store the External Data, which must contain ext\_id (the primary key), time, and at least one attribute.

\begin{figure}[t]
    \centering
    \includegraphics[width=0.95\columnwidth]{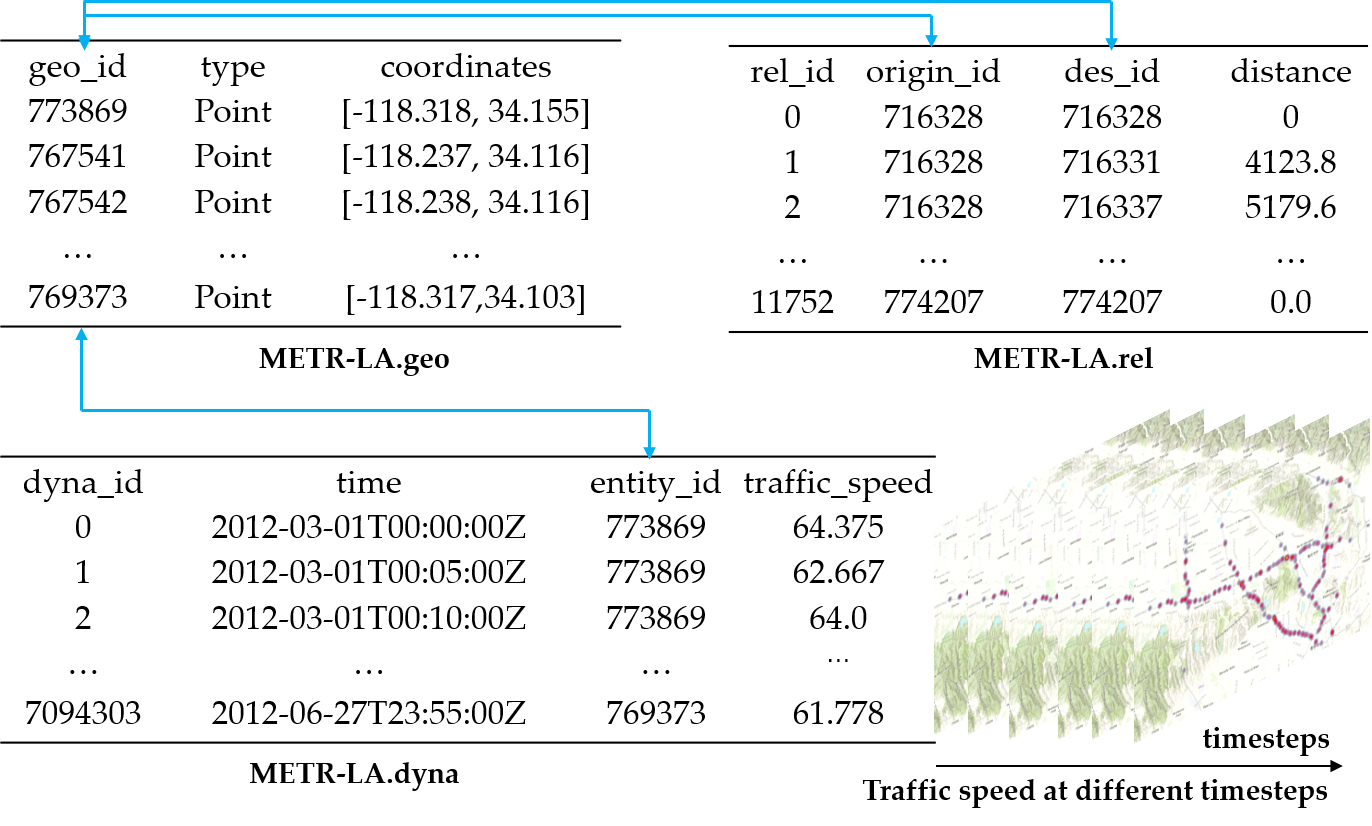}
    \vspace{-0.4cm}
    \caption{Example of atomic files: METR-LA dataset.}
    \vspace{-0.4cm}
    \label{fig:metr}
\end{figure}

To facilitate comprehension, we present an example of the Los Angeles traffic speed dataset in Figure~\ref{fig:metr}. This dataset comprises three atomic files: geo, rel, and dyna, as it does not include any corresponding external data. Notably, the origin\_id column, des\_id column within the ".rel" table, and the entity\_id column within the ".dyna" table are derived from the geo\_id column in the ".geo" table. Specifically, the ".geo" table contains the locations of 207 road sensors. The ".rel" table stores the distance relationships between these sensors, commonly used to construct graph structures. Lastly, the ".dyna" table records the average traffic speed information collected by these sensors at 5-minute intervals from March to June 2012. 

To validate the effectiveness of the atomic file format, we gathered 40 spatial-temporal dynamic datasets from 17 cities and transformed them into the atomic file format as presented in Table~\ref{tab:tsdataset}. The processed datasets, along with the associated data conversion scripts, are available at GitHub~\footnotemark[1]. We firmly believe that employing a unified data storage format can significantly alleviate the challenges associated with managing the urban spatial-temporal data. This standardization will enable more individuals to utilize urban spatial-temporal data and propel the rapid advancement of the urban spatial-temporal data field.

\footnotetext[1]{\url{https://github.com/LibCity/Bigscity-LibCity-Datasets}}

% Table generated by Excel2LaTeX from sheet '数据集总结'
\begin{table*}[hbtp]
  \centering
  \caption{Summary of Processed Urban Spatial-temporal Dynamic Datasets}
  \resizebox{\textwidth}{!}{
    \begin{tabular}{cccccccp{9em}}
    \toprule
    {\textbf{DATASET}} & 
    {\textbf{\#GEO}} & 
    {\textbf{\#REL}} & 
    {\textbf{\#DYNA}} & 
    {\textbf{PLACE}} & 
    {\textbf{DURATION}} & 
    {\textbf{\#TS}} & 
    \textbf{DATA TYPE} \\
    \midrule
    METR-LA~\cite{DCRNN} & 207   & 11,753  & 7,094,304  & Los Angeles, USA & Mar. 1, 2012 - Jun. 27, 2012 & 5min  & Graph Speed \\
    \midrule
    Los-Loop~\cite{TGCN} & 207   & 42,849  & 7,094,304  & Los Angeles, USA & Mar. 1, 2012 - Jun. 27, 2012 & 5min  & Graph Speed \\
    \midrule
    SZ-Taxi~\cite{TGCN} & 156   & 24,336  & 464,256  & Shenzhen, China & Jan. 1, 2015 - Jan. 31, 2015 & 15min & Graph Speed \\
    \midrule
    Q-Traffic~\cite{bbliaojqZhangKDD18deep} & 45,148  & 63,422  & 264,386,688  & Beijing, China & Apr. 1, 2017 - May 31, 2017 & 15min & Graph Speed \\
    \midrule
    Loop Seattle~\cite{cui2018deep, TGCLSTM} & 323   & 104,329  & 33,953,760  & Greater Seattle Area, USA & Jan. 1, 2015 - Dec. 31, 2015 & 5min  & Graph Speed \\
    \midrule
    PEMSD7(M)~\cite{STGCN} & 228   & 51,984  & 2,889,216  & California, USA & Weekdays of May. Jun., 2012 & 5min  & Graph Speed \\
    \midrule
    PEMS-BAY~\cite{DCRNN} & 325   & 8,358  & 16,937,700  & San Francisco Bay Area, USA & Jan. 1, 2017 - Jun. 30, 2017 & 5min  & Graph Speed \\
    \midrule
    Rotterdam~\cite{guopeng2020dynamic}  & 208   & - & 4,813,536  & Rotterdam, Holland & 135 days of 2018 & 2min  & Graph Speed \\
    \midrule
    PEMSD3~\cite{STSGCN} & 358   & 547   & 9,382,464  & California, USA & Sept. 1, 2018 - Nov. 30, 2018 & 5min  & Graph Flow \\
    \midrule
    PEMSD7~\cite{STSGCN} & 883   & 866   & 24,921,792  & California, USA & Jul. 1, 2016 - Aug. 31, 2016 & 5min  & Graph Flow \\
    \midrule
    Beijing subway~\cite{ResLSTM} & 276   & 76,176  & 248,400  & Beijing, China & Feb. 29, 2016 - Apr. 3, 2016 & 30min & Graph Flow \\
    \midrule
    M-dense~\cite{de2020spatio} & 30    & - & 525,600  & Madrid, Spain & Jan. 1, 2018 - Dec. 21, 2019 & 60min & Graph Flow \\
    \midrule
    SHMetro~\cite{liu2020physical} & 288   & 82,944  & 1,934,208  & Shanghai, China & Jul. 1, 2016 - Sept. 30, 2016 & 15min & Graph Flow \\
    \midrule
    HZMetro~\cite{liu2020physical} & 80    & 6,400  & 146,000  & Hangzhou, China & Jan. 1, 2019 - Jan. 25, 2019 & 15min & Graph Flow \\
    \midrule
    NYCTaxi-Dyna~\footnotemark[2] & 263   & 69,169  & 574,392  & New York, USA & Jan. 1, 2020 - Mar. 31, 2020 & 60min & Region Flow \\
    \midrule
    PEMSD4~\cite{ASTGCN} & 307   & 340   & 5,216,544  & San Francisco Bay Area, USA & Jan. 1, 2018 - Feb. 28, 2018 & 5min  & Graph Flow, Speed, Occupancy \\
    \midrule
    PEMSD8~\cite{ASTGCN} & 170   & 277   & 3,035,520  & San Bernardino Area, USA & Jul. 1, 2016 - Aug. 31, 2016 & 5min  & Graph Flow, Speed, Occupancy \\
    \midrule
    TaxiBJ2013~\cite{STResNet} & {32*32} & - & 4,964,352  & Beijing, China & Jul. 1, 2013 - Oct. 30, 2013 & 30min & Grid In\&Out Flow \\
    \midrule
    TaxiBJ2014~\cite{STResNet} & {32*32} & - & 4,472,832  & Beijing, China & Mar. 1, 2014 - Jun. 30, 2014 & 30min & Grid In\&Out Flow \\
    \midrule
    TaxiBJ2015~\cite{STResNet} & {32*32} & - & 5,652,480  & Beijing, China & Mar. 1, 2015 - Jun. 30, 2015 & 30min & Grid In\&Out Flow \\
    \midrule
    TaxiBJ2016~\cite{STResNet} & {32*32} & - & 6,782,976  & Beijing, China & Nov. 1, 2015 - Apr. 10, 2016 & 30min & Grid In\&Out Flow \\
    \midrule
    T-Drive~\cite{yuan2011driving, yuan2010t} & {32*32} & -  & 3,686,400  & Beijing, China & Feb. 1, 2015 - Jun. 30, 2015 & 60min & Grid In\&Out Flow \\
    \midrule
    Porto~\footnotemark[3] & {20*10} & - & 441,600  & Porto, Portugal & Jul. 1, 2013 - Sept. 30, 2013 & 60min & Grid In\&Out Flow \\
    \midrule
    NYCTaxi140103~\footnotemark[2] & {10*20} & - & 432,000  & New York, USA & Jan. 1, 2014 - Mar. 31, 2014 & 60min & Grid In\&Out Flow \\
    \midrule
    NYCTaxi140112~\cite{ACFM} & {15*5} & - & 1,314,000  & New York, USA & Jan. 1, 2014 - Dec. 31, 2014 & 30min & Grid In\&Out Flow \\
    \midrule
    NYCTaxi150103~\cite{STDN} & {10*20} & - & 576,000  & New York, USA & Jan. 1, 2015 - Mar. 1, 2015 & 30min & Grid In\&Out Flow \\
    \midrule
    NYCTaxi160102~\cite{DSAN} & {16*12} & - & 552,960  & New York, USA & Jan. 1, 2016 - Feb. 29, 2016 & 30min & Grid In\&Out Flow \\
    \midrule
    NYCBike140409~\cite{STResNet} & {16*8} & - & 562,176  & New York, USA & Apr. 1, 2014 - Sept. 30, 2014 & 60min & Grid In\&Out Flow \\
    \midrule
    NYCBike160708~\cite{STDN} & {10*20} & - & 576,000  & New York, USA & Jul. 1, 2016 - Aug. 29, 2016 & 30min & Grid In\&Out Flow \\
    \midrule
    NYCBike160809~\cite{DSAN} & {14*8} & - & 322,560  & New York, USA & Aug. 1, 2016 - Sept. 29, 2016 & 30min & Grid In\&Out Flow \\
    \midrule
    NYCBike200709~\footnotemark[4] & {10*20} & - & 441,600  & New York, USA & Jul. 1, 2020 - Sept. 30, 2020 & 60min & Grid In\&Out Flow \\
    \midrule
    AustinRide~\footnotemark[5] & {16*8} & - & 282,624  & Austin, USA & Jul. 1, 2016 - Sept. 30, 2016 & 60min & Grid In\&Out Flow \\
    \midrule
    BikeDC~\footnotemark[6] & {16*8} & - & 282,624  & Washington, USA & Jul. 1, 2020 - Sept. 30, 2020 & 60min & Grid In\&Out Flow \\
    \midrule
    BikeCHI~\footnotemark[7] & {15*18} & - & 596,160  & Chicago, USA & Jul. 1, 2020 - Sept. 30, 2020 & 60min & Grid In\&Out Flow \\
    \midrule
    NYCTaxi-OD~\footnotemark[2] & 263   & 69,169  & 150,995,927  & New York, USA & Apr. 1, 2020 - Jun. 30, 2020 & 60min & OD Flow \\
    \midrule
    NYC-TOD~\cite{CSTN} & {15*5} & - & 98,550,000  & New York, USA & Jan. 1, 2014 - Dec. 31, 2014     & 30min & Grid-OD Flow \\
    \midrule
    NYCTaxi150103~\cite{STDN} & {10*20} & - & 115,200,000  & New York, USA & Jan. 1, 2015 - Mar. 1, 2015 & 30min & Grid-OD Flow \\
    \midrule
    NYCBike160708~\cite{STDN} & {10*20} & - & 115,200,000  & New York, USA & Jul. 1, 2016 - Aug. 29, 2016 & 30min & Grid-OD Flow \\
    \midrule
    NYC-Risk~\cite{GSNet} & 243   & 59,049  & 3,504,000  & New York, USA & Jan. 1, 2013 - Dec. 31, 2013 & 60min & Risk \\
    \midrule
    CHI-Risk~\cite{GSNet} & 243   & 59,049  & 3,504,000  & New York, USA & Jan. 1, 2013 - Dec. 31, 2013 & 60min & Risk \\
    \bottomrule
    \end{tabular}
    }
  \label{tab:tsdataset}%
\end{table*}%
\footnotetext[2]{\url{https://www1.nyc.gov/site/tlc/about/tlc-trip-record-data.page}}
\footnotetext[3]{\url{https://www.kaggle.com/c/pkdd-15-predict-taxi-service-trajectory-i}}
\footnotetext[4]{\url{https://www.citibikenyc.com/system-data}}
\footnotetext[5]{\url{https://data.world/ride-austin/ride-austin-june-6-april-13}}
\footnotetext[6]{\url{https://www.capitalbikeshare.com/system-data}}
\footnotetext[7]{\url{https://www.divvybikes.com/system-data}}

\section{Problem Formalization} \label{tasks}
% \balance

The objective of urban spatial-temporal prediction tasks is to forecast future attributes based on historical observations of urban spatial-temporal data. The prediction tasks encompass various domains, such as traffic state prediction (\eg flow, speed), on-demand service prediction, air quality prediction, climate prediction (\eg wind speed, temperature), crime frequency prediction~\cite{stsurvey}.

In a formal sense, let us consider the urban spatial-temporal dynamic data $\boldsymbol{X}$ as defined in Section~\ref{stdynamicdata}. This problem aims to learn a mapping function $f$ that utilizes the previous $T$ time steps' observation values to predict the attributes of future $T'$ time steps~\cite{trafficsurvey1, trafficsurvey2, libcity}. The urban spatial-temporal prediction problem can be formulated as follows:
\begin{equation}\label{eq:problem_def}
[\boldsymbol{X}{(t-T+1)}, \cdots, \boldsymbol{X}t; \mathcal{G}] \stackrel{f}{\longrightarrow} [{\boldsymbol{X}}{(t+1)}, \cdots, {\boldsymbol{X}}{(t+T')}].
\end{equation}

\section{RoadMap for Spatial-Temporal Prediction Models} \label{model}
% \balance

% \subsection{Macro Group Prediction Tasks} \label{macro_models}

In this section, we present an overview of the technical development roadmap for urban spatial-temporal prediction models. The urban spatial-temporal prediction task aims to make forecasts by capturing the spatial-temporal dependencies in spatial and temporal data. Traditional statistical methods and machine learning techniques, such as VAR~\cite{VAR}, ARIMA~\cite{ARIMA}, and SVR~\cite{SVR}, rely on linear time-series analysis and fall short in effectively capturing spatial-temporal dependencies. However, the emergence of deep learning has opened up new possibilities for researchers to harness its potential in urban spatial-temporal prediction. Deep learning methods offer the ability to fully exploit the spatial-temporal dependencies in the data. The subsequent section presents a comprehensive summary of techniques for modeling spatial dependencies, temporal dependencies and fusing spatial-temporal information, which serves as the foundation for our experiments.

\subsection{Spatial Dependencies}
Spatial dependencies are present in urban spatial-temporal data, as per the first law of geography, that "everything is related to everything else, but things that are close to each other are more closely related." Moreover, due to the division of urban structural and functional areas, urban spatial-temporal data has a spatial hierarchy. Therefore, even distant locations may exhibit strong spatial dependencies due to the similarity of functions. The spatial dependencies in data are the core difference between spatial-temporal prediction tasks and the time series prediction tasks.

(1) \textbf{Convolutional Neural Networks (CNNs)}: CNNs are widely used for modeling Grid Relation Data as these data are Euclidean data. Researchers often divide cities into grids based on latitude and longitude, which allows the spatial-temporal attribute data to be viewed as images. In this way, CNNs can extract spatial dependencies from different grids~\cite{ECRNN, ACFM, DSAN, DMVSTNet}. For instance, ST-ResNet~\cite{STResNet} utilizes a deep residual CNN network for traffic flow prediction.

(2) \textbf{Graph Neural Networks (GNNs)}: GNNs are widely used for modeling Graph Relation Data due to their strong ability to represent graphs~\cite{DCRNN, STGCN, TGCN, STPEF, STSGCN, MRABGCN, SLCNN, DGCNN, MTGNN, AGCRN, STGDN, CCRNN, FCGAGA, DMSTGCN, STG-NCDE, MSDR, STDEN, CMOD, D2STGNN, METRO, MDTP}. Among GNNs, graph convolutional neural networks (GCNs)\cite{ChebConv} are the most commonly used for spatial-temporal prediction tasks. An essential issue in graph convolution models is the construction of the graph adjacency matrix, and early studies typically used static adjacency matrices~\cite{DCRNN, STGCN}. However, predefined graph adjacency matrices may not accurately reflect the actual spatial dependencies in spatial-temporal data. To address this, adaptive graph generation modules have been proposed~\cite{GWNET, AGCRN}. Additionally, some studies have focused on learning dynamic spatial dependencies to model the effects of changes in the data over time~\cite{DMSTGCN, ASTGNN, SLCNN}. To model continuous spatial dependencies, Graph ODE-based models have been used for spatial-temporal prediction~\cite{STGODE}. % For example,  GWNET~\cite{GWNET} employs graph convolution on a spatial dependency matrix. STFGNN~\cite{STFGNN} captures spatial-temporal dependencies synchronously through a spatial-temporal fusion graph module. STGODE~\cite{STGODE} takes attempts to construct a deeper GNN-based network for better mining spatial-temporal dynamics.

(3) \textbf{Spatial Attention Mechanisms}: The spatial attention mechanism plays an essential role in modeling time-varying spatial dependencies between geographical entities~\cite{ACFM, ASTGNN, GaAN, GMAN, ASTGCN, STAGGCN, STTN, TFormer, MGT, PDFormer}. Through the spatial attention, models can autonomously learn the dynamic spatial dependencies embedded in the spatial-temporal data by assigning adaptive weights to different locations at different time steps. For instance, GMAN~\cite{GMAN} proposes both spatial and temporal attention mechanisms and designs a gated fusion to fuse spatial and temporal information. ASTGNN~\cite{ASTGNN} introduces a dynamic graph convolution module with a self-attention mechanism to capture spatial dependencies in a dynamic graph.

\subsection{Temporal Dependencies} \label{temporaldepend}
Urban spatial-temporal data displays temporal dependencies, similar to time-series data, which exhibit characteristics such as closeness, periodicity, and trend. For example, a traffic jam at 8 am can impact the traffic situation at 9 am, and rush hours tend to occur at similar times on consecutive workdays following a 24-hour cycle. Furthermore, holidays and weekdays can also have distinct effects on urban spatial-temporal data patterns.

(1) \textbf{Recurrent Neural Networks (RNNs)}: RNNs are a type of neural network designed to process sequential data and are thus well-suited for capturing the temporal dependencies in spatial-temporal data~\cite{ACFM, DMVSTNet, DCRNN, AGCRN, CCRNN, TGCN, TGCLSTM, TFormer, MDTP}. To address the RNNs' inability to model long-term dependencies, LSTM~\cite{LSTM} and GRU~\cite{GRU} are introduced as variants of the original RNN model. These models incorporate a gating mechanism that allows them to capture longer-term dependencies. The Seq2Seq model~\cite{Seq2Seq} is a popular framework that uses RNNs as both encoders and decoders to enable multi-step spatial-temporal prediction.

(2) \textbf{Temporal Convolutional Networks (TCNs)}: TCNs are designed to process sequential data in a parallel fashion, which is impossible with Recurrent Neural Networks (RNNs) due to their reliance on historical information. TCNs are essentially 1-D CNN models that are used in the field of time series forecasting. They consider only the information before a given time step and are called causal convolutions. To increase the receptive field, dilated convolutions are used with causal convolutions. Multiple layers of dilated causal convolutions (also known as WaveNet~\cite{WaveNet}) can be stacked to achieve exponential growth of the receptive field. TCNs / 1D-CNN have become a common infrastructure in the field of spatial-temporal prediction~\cite{GWNET, MTGNN, STAGGCN, ASTGCN, STGCN, HGCN, DGCN}.

(3) \textbf{Temporal Attention Mechanisms}: The temporal attention mechanism enables adaptive learning of nonlinear temporal dependencies in spatial-temporal data by assigning different weights to data at different time steps~\cite{ASTGCN, GMAN, DSAN, DGCN, STAGGCN, STG2Seq, STTN, PDFormer}. This mechanism helps address the limitation of RNN models in modeling long-term dependencies, especially when combined with the Seq2Seq model. Recently, self-attention-based Transformer models have shown promising results in time series forecasting and have been applied to urban spatial-temporal prediction. For example, PDFormer~\cite{PDFormer} introduces a spatial-temporal self-attention mechanism to urban traffic flow prediction and incorporates the time-delay property of traffic state propagation.

\subsection{Spatial-Temporal Dependencies Fusion}
To capture the spatial-temporal dependencies in data, it is common to combine spatial and temporal models into a hybrid model, such as CNN+RNN, GCN+RNN, GCN+TCN, etc. The following are some common methods for fusing spatial and temporal models:

(1) \textbf{Sequential Structure}: The sequential structure combines the spatial and temporal neural networks in either parallel or serial configurations to form a spatial-temporal block~\cite{STGCN, GWNET, MTGNN, TGCN, TGCLSTM, ToGCN, ASTGCN, GMAN, STTN}. For example, STGCN~\cite{STGCN} utilizes 2 TCNs and 1 GCN to create a spatial-temporal block and stack them serially to capture the spatial-temporal dependencies in the data. TGCN~\cite{TGCN} adopts a recursive approach by combining GCNs and GRUs, processing the input data recursively with GRUs after passing through GCNs at each time step. PDFormer~\cite{PDFormer} employs multiple spatial self-attention heads and temporal self-attention heads to learn the spatial-temporal dependencies in the data.

(2) \textbf{Coupled Structure}: The coupled structure integrates spatial neural networks into the computation of temporal neural networks, most commonly by combining GCNs and RNNs~\cite{DCRNN, CCRNN, MRABGCN, AGCRN}. In this approach, the fully connected operations in the computation of RNNs, including variants like GRU and LSTM models, are replaced with GCN operations. By doing so, the models can incorporate spatial dependencies into the learning of temporal dependencies. Some examples of this approach include DCRNN~\cite{DCRNN}, CCRNN~\cite{CCRNN}, MRABGCN~\cite{MRABGCN}, and AGCRN~\cite{AGCRN}, etc.

(3) \textbf{Spatial-Temporal Synchronized Learning}: Previous research on modeling spatial-temporal dependencies in data has often used separate components to independently learn spatial and temporal dependencies. However, a spatial-temporal synchronous learning approach can simultaneously directly capture local spatial-temporal dependencies in the data. To accomplish this, a spatial-temporal local graph structure is constructed across time steps, and GCN is used to learn spatial-temporal dependencies. STSGCN~\cite{STSGCN} and STFGCN~\cite{STFGNN} are two examples of this approach.

\section{Experiments} \label{experiments}
% \balance

In this section, we perform extensive experiments using diverse models and datasets with the aim of establishing a performance leaderboard. Our objective is to compare previous works in the field of spatial-temporal prediction and identify noteworthy directions for future research. Through these experiments, we seek to provide valuable insights and recommendations.

% We have conducted comprehensive experiments on various models and datasets to demonstrate the superiority of our proposed library in facilitating a fair comparison of commonly used models.

% \subsection{Benchmarks on Macro Group Prediction Tasks} 

% The Macro Group Prediction Tasks involve predicting the spatial-temporal attributes of macro groups in the future based on historical observations of Spatial Static Temporal Dynamic (SSTD) data. Since the technical development lines for different macro group prediction tasks are similar, only the datasets' attributes vary. We conduct a unified experiment on different datasets to compare the performance of different model structures.

\subsection{Datasets} 
In this section, we consider a total of 20 datasets, categorized into two types: Graph Relation Data and Grid Relation Data. The datasets we use are as follows. New datasets released in this study do not have any citations followed, while others with citations are open-sourced datasets. Please refer to Table~\ref{tab:tsdataset} for more details.

(1) Graph Relation Data: These datasets consist of traffic speed or flow data recorded by traffic sensors. The following datasets are considered: METR-LA~\cite{DCRNN}, PEMS-BAY~\cite{DCRNN}, PEMSD7(M)~\cite{STGCN}, PEMSD3~\cite{ASTGCN}, PEMSD4-Speed~\cite{ASTGCN}, PEMSD8-Speed~\cite{ASTGCN}, PEMSD4-Flow~\cite{ASTGCN}, PEMSD8-Flow~\cite{ASTGCN}, PEMSD4-Occupy~\cite{ASTGCN}, and PEMSD8-Occupy~\cite{ASTGCN}.

(2) Grid Relation Data: These datasets consist of city grid traffic inflow and outflow data based on the urban taxi or bike trajectories. The following datasets are considered: NYCTaxi150103~\cite{STDN}, NYCTaxi160102~\cite{DSAN}, NYCTaxi140103*, TaxiBJ2014~\cite{STResNet}, TaxiBJ2015~\cite{STResNet}, NYCBike160708~\cite{STDN}, NYCBike140409~\cite{STResNet}, BikeDC*, BikeCHI*, and NYCBike200709*.

\subsection{Models} 

We conducted experiments on the following 18 models that can be divided into six categories as follows:

(1) \textbf{General Time Series Prediction Models}
% (1) General Macro Group Prediction Baseline Methods:
\begin{itemize}
    \item Seq2Seq~\cite{Seq2Seq}: uses the encode-decoder framework based on the gated recurrent unit.
    \item AutoEncoder~\cite{AutoEncoder}: uses an encoder to learn the embedding and then a decoder to predict the future.
    \item FNN~\cite{DCRNN}: Feed forward neural network with two hidden layers and L2 regularization.
\end{itemize}

(2) \textbf{Sequential Structure Models (Spatial-CNNs plus others)}
\begin{itemize}
    \item {STResNet}~\cite{STResNet} models the spatial-temporal correlations by residual unit.
    \item {ACFM}~\cite{ACFM}: infers the evolution of the crowd flow with a ConvLSTM and attention mechanism
    \item {DMVSTNet}~\cite{DMVSTNet} combines CNN and LSTM to capture the spatial, temporal, and semantic views.
    % \item {DSAN}~\cite{DSAN} models the spatial-temporal correlations through a Multi-Space Attention mechanism.
\end{itemize}

(3) \textbf{Sequential Structure Models (Spatial-GCNs plus others)}
\begin{itemize}
    % GNN + CNN
    \item {STGCN}~\cite{STGCN} combines GCNs and gated temporal convolution.
    % GNN + casual TCN
    \item {GWNET}~\cite{GWNET} combines adaptive adjacency matrix into GCNs and 1D dilated casual convolutions.
    \item {MTGNN}~\cite{MTGNN} combines adaptive GCNs with mix-hop propagation layers and dilated inception layers.
    % GNN + RNN
    \item {TGCN}~\cite{TGCN} combines GCNs and the gated recurrent unit.
    % \item {TGCLSTM}~\cite{TGCLSTM} combines the K-hop graph convolutional network and long short-term memory network.
    % \item {ToGCN}~\cite{ToGCN} combines graph convolutional network and LSTM-based encoder-decoder framework.
\end{itemize}
    
(4) \textbf{Sequential Structure Models (Spatial-Attentions plus others)}
\begin{itemize}
    \item {ASTGCN}~\cite{ASTGCN} pluses the STGCN with spatial-temporal attention mechanism.
    \item {GMAN}~\cite{GMAN} consists of multiple spatial-temporal attention blocks in the encoder-decoder framework.
    \item {STTN}~\cite{STTN} combines the spatial and temporal transformer and graph convolutional network.
\end{itemize}

(5) \textbf{Coupled Structure Models (GCNs + RNNs)}
\begin{itemize}
    \item DCRNN~\cite{DCRNN}: couples the diffusion GCN and the encoder-decoder framework.
    \item AGCRN~\cite{AGCRN}: couples the adaptive GCN and the gated recurrent unit.
    % \item CCRNN~\cite{CCRNN}: couples the coupled layer-wise graph convolution and the encoder-decoder framework.
\end{itemize}

(6) \textbf{Spatial-Temporal Synchronized Learning Models}
\begin{itemize}
    \item {STG2Seq}~\cite{STG2Seq} utilizes a hierarchical graph convolutional to capture spatial and temporal correlations simultaneously.
    \item {STSGCN}~\cite{STSGCN} utilizes spatial-temporal synchronous modeling mechanism to model localized correlations.
    \item {D2STGNN}~\cite{D2STGNN} utilizes the diffusion and inherent modules to model the diffusion process and the inherent signal.
    % \item STFGNN~\cite{STFGNN} designs a spatial-temporal fusion graph module to capture spatial-temporal correlations synchronously.
\end{itemize}

% (7) \textbf{Continuous Spatial-Temporal Learning Models}
% \begin{itemize}
%     \item STGODE~\cite{STGODE} proposed a spatial-temporal graph ordinary differential equation for continuous spatial-temporal modeling.
% \end{itemize}

\subsection{Experimental Settings} 

\subsubsection{Environment Settings} All experiments are conducted using NVIDIA 3090 GPUs and 256 GB memory on Ubuntu 20.04. We complete the implementation of the models with the help of the LibCity~\cite{libcity} framework, using Python 3.9.7 and PyTorch 1.10.1~\cite{pytorch}.

\subsubsection{Dataset Processing} The datasets are split into training, validation, and test sets with a ratio of \textbf{7:1:2}. For graph relation Data, we perform multi-step prediction using 12 past steps to predict 12 future steps. For grid relation Data, we perform single-step prediction using six past steps to predict the traffic inflow and outflow of the next single step. % Before training, Z-score normalization is applied to standardize the inputs.

\subsubsection{Model and Training Settings} The Adam optimizer is used with a batch size of 64 (adjusted as necessary when OOM occurred for some large models), a training epoch of 100, and an early stopping mechanism when the valid set loss lasts 30 epochs without dropping. Gradient clipping with a gradient clipping value of 5 is employed. The model parameters are set according to the settings in the original paper. Additionally, the time of day is introduced as auxiliary data to enhance prediction accuracy, following~\cite{GWNET}. To ensure a fair comparison, only the recent components of the ASTGCN~\cite{ASTGCN} and ST-ResNet~\cite{STResNet} models are utilized.

\subsubsection{Evaluation Metrics} Three widely-used metrics are employed to evaluate the models: (1) Mean Absolute Error (MAE), (2) Mean Absolute Percentage Error (MAPE), and (3) Root Mean Squared Error (RMSE). Missing values are excluded from training and testing, following~\cite{DCRNN}. For grid relation datasets, samples with flow values below five are filtered, consistent with~\cite{DMVSTNet}. Due to spatial constraints, for graph relation data, the reported results are the average of multi-step predictions, while for grid relation data, the reported results are the average of inflows and outflows. We repeat all experiments five times and report the average results.

\begin{table}[t]
  \centering
  \small
  \caption{Model Performance Leaderboard}
  \resizebox{\columnwidth}{!}{
    \begin{tabular}{c|c|c|c|c|c}
    \toprule
    1     & 2     & 3     & 4     & 5     & 6 \\
    \midrule
    D2STGNN & MTGNN & GWNET & AGCRN  & GMAN & DCRNN \\
    \midrule
    7     & 8     & 9     & 10    & 11    & 12 \\
    \midrule
    STGCN & STTN  & STSGCN & STG2Seq & ASTGCN & TGCN \\
    \midrule
    13    & 14    & 15    & 16    & 17    & 18 \\
    \midrule
    DMVSTNet & ACFM  & STResNet & Seq2Seq & FNN   & AutoEncoder \\
    \bottomrule
    \end{tabular}%
    }
    \vspace{-0.2cm}
  \label{tab:rank}%
  \vspace{-0.3cm}
\end{table}%

% Table generated by Excel2LaTeX from sheet 'Sheet4'
\begin{table*}[t]
  \centering
  \caption{Performance Comparision on Graph Relation Data}
  \resizebox{\textwidth}{!}{
    \begin{tabular}{c|ccc|ccc|ccc|ccc|ccc|c}
    \toprule
    Dataset & \multicolumn{3}{c|}{METR-LA} & \multicolumn{3}{c|}{PEMS-BAY} & \multicolumn{3}{c|}{PEMSD7(M)} & \multicolumn{3}{c|}{PEMSD4-Speed} & \multicolumn{3}{c|}{PEMSD8-Speed} &  \\
    \midrule
    Metric & MAE   & MAPE  & RMSE  & MAE   & MAPE  & RMSE  & MAE   & MAPE  & RMSE  & MAE   & MAPE  & RMSE  & MAE   & MAPE  & RMSE  & RANK \\
    \midrule
    Seq2Seq & 4.65  & 11.39  & 7.79  & 2.57  & 6.07  & 5.32  & 8.92  & 21.90  & 11.47  & 2.26  & 5.08  & 4.88  & 1.91  & 4.62  & 4.63  & 13 \\
    AutoEncoder & 4.11  & 12.01  & 7.76  & 2.52  & 6.13  & 5.20  & 8.97  & 20.18  & 11.65  & 2.33  & 5.45  & 5.14  & 2.17  & 5.22  & 5.22  & 15 \\
    FNN   & 4.35  & 12.73  & 8.37  & 2.05  & 6.16  & 4.69  & 8.82  & 20.16  & 11.46  & 1.98  & 5.36  & 5.17  & 1.89  & 4.72  & 5.08  & 14 \\
    \midrule
    STGCN & 3.29  & 8.80  & 6.59  & 1.71  & 3.83  & 3.91  & 8.44  & 18.71  & 9.99  & 1.71  & 3.64  & 3.72  & 1.42  & 3.32  & 3.42  & 8 \\
    GWNET & 3.07  & 8.32  & 6.16  & 1.58  & 3.56  & 3.55  & 8.07  & 18.56  & 10.39  & 1.57  & 3.32  & 3.60  & 1.29  & 2.88  & 3.23  & 2 \\
    MTGNN & 3.02  & 8.25  & 6.09  & 1.59  & 3.59  & 3.58  & 8.02  & 18.33  & 9.86  & 1.60  & 3.42  & 3.69  & 1.29  & 2.89  & 3.21  & 3 \\
    TGCN  & 3.52  & 9.93  & 6.87  & 1.82  & 4.24  & 3.85  & 8.74  & 18.91  & 11.37  & 1.90  & 3.80  & 3.75  & 1.73  & 3.79  & 3.50  & 12 \\
    \midrule
    ASTGCN & 3.47  & 9.82  & 6.73  & 1.90  & 4.28  & 4.25  & 8.50  & 19.51  & 11.33  & 1.79  & 3.82  & 3.74  & 1.49  & 3.32  & 3.49  & 11 \\
    GMAN  & 3.16  & 8.65  & 6.43  & 1.57  & 3.65  & 3.64  & 7.67  & 18.04  & 9.76  & 1.68  & 3.65  & 3.63  & 1.35  & 3.13  & 3.21  & 5 \\
    STTN  & 3.26  & 9.07  & 6.54  & 1.68  & 3.81  & 3.85  & 8.70  & 18.99  & 10.87  & 1.64  & 3.45  & 3.68  & 1.42  & 3.47  & 3.68  & 7 \\
    \midrule
    DCRNN & 3.16  & 8.66  & 6.44  & 1.67  & 3.80  & 3.83  & 8.52  & 19.58  & 10.46  & 1.65  & 3.46  & 3.76  & 1.40  & 3.10  & 3.43  & 6 \\
    AGCRN & 3.17  & 8.79  & 6.35  & 1.62  & 3.68  & 3.63  & 7.76  & 18.19  & 10.39  & 1.62  & 3.41  & 3.64  & 1.34  & 3.01  & 3.27  & 4 \\
    \midrule
    STG2Seq & 3.41  & 9.33  & 6.64  & 1.76  & 3.84  & 3.94  & 8.43  & 19.85  & 10.53  & 1.85  & 3.60  & 3.79  & 1.59  & 3.58  & 3.74  & 10 \\
    STSGCN & 3.34  & 9.26  & 6.65  & 1.75  & 3.81  & 3.90  & 8.08  & 19.55  & 10.98  & 1.84  & 3.55  & 3.73  & 1.53  & 3.51  & 3.73  & 9 \\
    D2STGNN & 2.91  & 7.93  & 5.84  & 1.56  & 3.59  & 3.46  & 2.49  & 6.26  & 5.03  & 1.55  & 3.28  & 3.50  & 1.28  & 3.07  & 3.32  & 1 \\
    \midrule
    \midrule
    Dataset & \multicolumn{3}{c|}{PEMSD4-Flow} & \multicolumn{3}{c|}{PEMSD8-Flow} & \multicolumn{3}{c|}{PEMSD4-Occupy} & \multicolumn{3}{c|}{PEMSD8-Occupy} & \multicolumn{3}{c|}{PEMSD3} &  \\
    \midrule
    Metric & MAE   & MAPE  & RMSE  & MAE   & MAPE  & RMSE  & MAE   & MAPE  & RMSE  & MAE   & MAPE  & RMSE  & MAE   & MAPE  & RMSE  & RANK \\
    \midrule
    Seq2Seq & 22.84  & 16.30  & 36.72  & 20.35  & 12.79  & 33.15  & 0.93  & 20.70  & 2.40  & 0.89  & 14.46  & 2.39  & 20.07  & 19.55  & 34.22  & 13 \\
    AutoEncoder & 24.42  & 17.83  & 38.38  & 21.32  & 13.73  & 35.11  & 0.98  & 21.25  & 2.49  & 0.99  & 15.91  & 2.57  & 23.12  & 23.36  & 38.36  & 15 \\
    FNN   & 24.17  & 17.73  & 39.48  & 20.74  & 13.21  & 32.03  & 0.94  & 21.42  & 2.29  & 0.83  & 15.41  & 2.18  & 20.50  & 19.59  & 33.20  & 14 \\
    \midrule
    STGCN & 20.39  & 13.95  & 32.41  & 15.63  & 10.50  & 24.63  & 0.77  & 16.52  & 2.08  & 0.71  & 11.54  & 1.89  & 15.81  & 15.31  & 27.10  & 7 \\
    GWNET & 18.64  & 13.36  & 30.03  & 14.51  & 9.39  & 23.36  & 0.73  & 16.32  & 1.95  & 0.65  & 10.68  & 1.85  & 14.50  & 13.97  & 24.98  & 1 \\
    MTGNN & 18.68  & 13.99  & 30.05  & 14.88  & 9.93  & 23.81  & 0.73  & 16.15  & 1.93  & 0.66  & 10.87  & 1.85  & 14.78  & 14.16  & 25.61  & 3 \\
    TGCN  & 20.91  & 14.80  & 33.38  & 16.26  & 11.86  & 25.65  & 0.84  & 17.71  & 2.15  & 0.77  & 12.14  & 1.94  & 16.98  & 18.62  & 28.26  & 12 \\
    \midrule
    ASTGCN & 20.37  & 14.24  & 32.73  & 16.36  & 11.89  & 24.95  & 0.83  & 17.67  & 2.17  & 0.75  & 11.95  & 1.92  & 16.80  & 16.35  & 27.71  & 10 \\
    GMAN  & 19.44  & 14.52  & 30.79  & 15.67  & 10.99  & 24.20  & 0.75  & 16.76  & 2.01  & 0.66  & 11.59  & 1.85  & 15.60  & 16.33  & 27.30  & 5 \\
    STTN  & 19.89  & 14.67  & 31.14  & 16.10  & 11.28  & 24.70  & 0.79  & 17.16  & 2.11  & 0.70  & 11.58  & 1.94  & 16.07  & 17.41  & 27.93  & 8 \\
    \midrule
    DCRNN & 19.97  & 13.87  & 31.53  & 15.45  & 9.86  & 24.11  & 0.75  & 16.85  & 2.03  & 0.69  & 11.38  & 1.87  & 15.66  & 16.39  & 27.92  & 6 \\
    AGCRN & 18.74  & 12.81  & 30.60  & 14.84  & 9.93  & 24.06  & 0.74  & 16.59  & 1.96  & 0.68  & 11.52  & 1.85  & 15.21  & 15.03  & 27.02  & 4 \\
    \midrule
    STG2Seq & 20.13  & 14.90  & 31.36  & 16.27  & 11.32  & 24.93  & 0.82  & 17.64  & 2.18  & 0.73  & 11.97  & 1.97  & 16.44  & 16.41  & 28.13  & 11 \\
    STSGCN & 20.15  & 14.60  & 31.27  & 16.16  & 11.29  & 24.82  & 0.81  & 17.40  & 2.17  & 0.73  & 11.91  & 1.95  & 16.36  & 16.37  & 28.11  & 9 \\
    D2STGNN & 18.47  & 13.04  & 30.19  & 14.67  & 9.67  & 23.50  & 0.73  & 16.24  & 1.93  & 0.67  & 11.17  & 1.81  & 14.66  & 14.70  & 25.81  & 2 \\
    \bottomrule
    \end{tabular}%
    }
    \vspace{-0.2cm}
  \label{tab:graph}%
  \vspace{-0.3cm}
\end{table*}%

% Table generated by Excel2LaTeX from sheet 'Sheet4'
\begin{table*}[t]
  \centering
  \caption{Performance Comparision on Grid Relation Data}
  \resizebox{\textwidth}{!}{
    \begin{tabular}{c|ccc|ccc|ccc|ccc|ccc|c}
    \toprule
    Dataset & \multicolumn{3}{c|}{NYCTaxi150103} & \multicolumn{3}{c|}{NYCTaxi160102} & \multicolumn{3}{c|}{NYCTaxi140103} & \multicolumn{3}{c|}{TaxiBJ2014} & \multicolumn{3}{c|}{TaxiBJ2015} &  \\
    \midrule
    Metric & MAE   & MAPE  & RMSE  & MAE   & MAPE  & RMSE  & MAE   & MAPE  & RMSE  & MAE   & MAPE  & RMSE  & MAE   & MAPE  & RMSE  & RANK \\
    \midrule
    Seq2Seq & 12.95  & 20.21  & 21.23  & 12.89  & 20.35  & 21.23  & 8.92  & 21.90  & 11.47  & 16.94  & 18.93  & 21.93  & 18.29  & 21.16  & 27.88  & 18 \\
    AutoEncoder & 12.56  & 20.86  & 21.90  & 12.83  & 20.11  & 22.37  & 8.97  & 20.18  & 11.65  & 15.00  & 17.31  & 21.05  & 18.53  & 22.57  & 22.84  & 17 \\
    FNN   & 14.12  & 22.32  & 21.68  & 13.65  & 22.22  & 19.84  & 8.82  & 20.16  & 11.46  & 14.43  & 17.52  & 21.74  & 15.99  & 18.69  & 21.62  & 16 \\
    \midrule
    STResNet & 12.12  & 20.18  & 20.71  & 11.96  & 18.81  & 16.59  & 9.09  & 19.96  & 11.84  & 11.82  & 15.64  & 18.83  & 12.85  & 16.87  & 18.93  & 15 \\
    ACFM  & 12.24  & 20.06  & 20.53  & 11.40  & 18.78  & 16.31  & 8.77  & 19.70  & 11.63  & 11.74  & 15.83  & 18.71  & 12.83  & 16.33  & 18.90  & 14 \\
    DMVSTNet & 11.99  & 19.97  & 20.47  & 11.53  & 18.61  & 16.23  & 8.75  & 19.66  & 11.40  & 11.71  & 15.64  & 18.67  & 12.67  & 15.24  & 18.75  & 13 \\
    \midrule
    STGCN & 10.84  & 18.48  & 18.64  & 10.68  & 18.58  & 14.95  & 8.44  & 18.71  & 9.99  & 11.47  & 14.46  & 18.63  & 11.62  & 14.58  & 17.34  & 7 \\
    GWNET & 10.24  & 17.37  & 16.71  & 10.48  & 17.98  & 15.28  & 8.07  & 18.56  & 10.39  & 11.43  & 13.98  & 18.34  & 11.22  & 14.71  & 18.38  & 5 \\
    MTGNN & 10.42  & 17.19  & 15.80  & 10.00  & 16.80  & 14.50  & 8.02  & 18.33  & 9.86  & 11.48  & 14.05  & 17.38  & 11.03  & 14.37  & 17.05  & 1 \\
    TGCN  & 11.97  & 18.93  & 20.27  & 10.11  & 18.16  & 14.86  & 8.74  & 18.91  & 11.37  & 11.62  & 15.59  & 18.18  & 11.87  & 15.11  & 18.72  & 12 \\
    \midrule
    ASTGCN & 11.97  & 19.99  & 20.32  & 10.48  & 18.19  & 14.83  & 8.50  & 19.51  & 11.33  & 11.44  & 14.35  & 18.33  & 12.27  & 15.27  & 17.63  & 11 \\
    GMAN  & 9.83  & 17.62  & 15.53  & 9.61  & 17.60  & 14.12  & 7.67  & 18.04  & 9.76  & 11.48  & 14.26  & 18.05  & 11.23  & 15.17  & 17.58  & 4 \\
    STTN  & 10.55  & 20.05  & 17.20  & 9.86  & 18.73  & 14.50  & 8.70  & 18.99  & 10.87  & 11.20  & 15.15  & 18.62  & 11.36  & 15.02  & 18.04  & 8 \\
    \midrule
    DCRNN & 10.00  & 17.81  & 14.86  & 9.82  & 17.53  & 14.00  & 8.52  & 19.58  & 10.46  & 11.68  & 14.70  & 17.44  & 11.36  & 15.11  & 17.56  & 6 \\
    AGCRN & 10.02  & 17.52  & 17.92  & 9.79  & 17.30  & 14.39  & 7.76  & 18.19  & 10.39  & 11.40  & 14.14  & 18.25  & 11.12  & 14.48  & 18.06  & 3 \\
    \midrule
    STG2Seq & 11.21  & 18.82  & 18.27  & 11.27  & 17.63  & 15.41  & 8.43  & 19.85  & 10.53  & 11.63  & 14.36  & 18.52  & 11.33  & 15.21  & 18.10  & 10 \\
    STSGCN & 10.93  & 18.95  & 16.94  & 10.28  & 17.74  & 15.24  & 8.08  & 19.55  & 10.98  & 11.69  & 14.38  & 18.53  & 11.56  & 15.11  & 18.45  & 9 \\
    D2STGNN & 10.24  & 17.53  & 16.53  & 10.05  & 17.33  & 14.41  & 7.67  & 17.15  & 9.83  & 11.44  & 14.17  & 17.62  & 11.15  & 14.52  & 17.53  & 2 \\
    \midrule
    \midrule
    Dataset & \multicolumn{3}{c|}{NYCBike160708} & \multicolumn{3}{c|}{NYCBike140409} & \multicolumn{3}{c|}{NYCBike200709} & \multicolumn{3}{c|}{BikeDC} & \multicolumn{3}{c|}{BikeCHI} &  \\
    \midrule
    Metric & MAE   & MAPE  & RMSE  & MAE   & MAPE  & RMSE  & MAE   & MAPE  & RMSE  & MAE   & MAPE  & RMSE  & MAE   & MAPE  & RMSE  & RANK \\
    \midrule
    Seq2Seq & 5.61  & 26.14  & 7.35  & 5.61  & 25.83  & 7.81  & 8.57  & 25.52  & 14.34  & 10.88  & 34.97  & 15.09  & 5.80  & 27.92  & 8.53  & 16 \\
    AutoEncoder & 5.81  & 26.79  & 7.55  & 5.63  & 25.72  & 7.89  & 8.59  & 25.54  & 14.58  & 10.94  & 32.57  & 14.83  & 5.88  & 28.72  & 8.68  & 17 \\
    FNN   & 5.80  & 27.12  & 7.90  & 6.15  & 25.59  & 7.77  & 8.62  & 25.43  & 14.72  & 10.87  & 32.77  & 14.99  & 6.12  & 29.38  & 8.78  & 18 \\
    \midrule
    STResNet & 5.53  & 25.85  & 7.05  & 5.86  & 26.63  & 7.97  & 8.86  & 25.40  & 14.18  & 10.77  & 30.38  & 14.89  & 5.62  & 27.19  & 7.96  & 15 \\
    ACFM  & 5.64  & 25.89  & 6.87  & 5.68  & 26.30  & 7.85  & 8.75  & 25.38  & 14.27  & 10.73  & 30.25  & 14.83  & 5.54  & 26.66  & 7.90  & 14 \\
    DMVSTNet & 5.32  & 25.80  & 6.76  & 5.63  & 26.29  & 7.75  & 8.70  & 25.32  & 14.08  & 10.67  & 30.15  & 14.70  & 5.53  & 26.41  & 7.71  & 13 \\
    \midrule
    STGCN & 4.64  & 23.21  & 6.53  & 5.31  & 24.96  & 6.91  & 7.52  & 22.30  & 11.31  & 10.10  & 27.13  & 13.66  & 4.67  & 24.84  & 6.42  & 7 \\
    GWNET & 4.55  & 23.22  & 6.48  & 5.19  & 24.19  & 6.85  & 7.47  & 22.38  & 11.11  & 10.14  & 26.30  & 12.88  & 4.72  & 24.61  & 6.24  & 5 \\
    MTGNN & 4.49  & 22.58  & 5.87  & 4.94  & 23.57  & 6.41  & 7.10  & 21.97  & 10.36  & 9.96  & 26.89  & 12.68  & 4.50  & 24.60  & 5.69  & 1 \\
    TGCN  & 4.82  & 25.81  & 6.51  & 5.46  & 24.43  & 7.07  & 8.55  & 25.34  & 13.74  & 10.06  & 29.68  & 14.63  & 5.51  & 26.40  & 6.77  & 12 \\
    \midrule
    ASTGCN & 5.22  & 24.53  & 6.68  & 5.57  & 24.13  & 7.44  & 8.27  & 25.21  & 13.82  & 10.16  & 29.04  & 14.34  & 5.42  & 26.41  & 5.87  & 11 \\
    GMAN  & 4.53  & 22.90  & 5.95  & 5.03  & 24.44  & 6.85  & 7.43  & 22.24  & 11.40  & 9.87  & 26.11  & 12.11  & 4.75  & 24.60  & 6.25  & 3 \\
    STTN  & 4.65  & 24.22  & 6.37  & 5.14  & 25.24  & 6.80  & 7.50  & 23.49  & 11.72  & 9.97  & 27.79  & 12.22  & 4.81  & 25.75  & 6.60  & 8 \\
    \midrule
    DCRNN & 4.50  & 23.13  & 6.38  & 5.12  & 24.47  & 6.93  & 7.44  & 22.43  & 11.42  & 9.95  & 27.11  & 13.26  & 4.73  & 25.62  & 6.21  & 6 \\
    AGCRN & 4.52  & 23.02  & 5.97  & 5.13  & 24.43  & 6.65  & 7.54  & 22.38  & 11.45  & 9.96  & 27.57  & 12.62  & 4.78  & 25.61  & 6.11  & 4 \\
    \midrule
    STG2Seq & 5.08  & 24.26  & 6.74  & 5.41  & 24.46  & 7.09  & 8.04  & 25.05  & 11.93  & 10.11  & 28.09  & 13.77  & 4.80  & 25.86  & 6.58  & 10 \\
    STSGCN & 4.63  & 23.66  & 6.64  & 5.40  & 24.67  & 7.18  & 7.62  & 23.16  & 11.81  & 9.97  & 27.61  & 12.99  & 4.81  & 26.21  & 6.44  & 9 \\
    D2STGNN & 4.65  & 22.75  & 6.12  & 4.98  & 23.50  & 6.44  & 7.30  & 22.86  & 10.64  & 9.30  & 25.16  & 11.69  & 4.68  & 25.35  & 5.96  & 2 \\
    \bottomrule
    \end{tabular}%
    }
    \vspace{-0.2cm}
  \label{tab:grid}%
  \vspace{-0.3cm}
\end{table*}%

\subsection{Performance Comparision}
Table~\ref{tab:rank} represents the model performance leaderboard for all models on all datasets. Tables~\ref{tab:graph} and Table~\ref{tab:grid} present the experiment results for graph relation data and grid relation data, respectively. From these results, we draw the following observations:

(1) General time series prediction models, including Seq2Seq, AutoEncoder, and FNN, perform poorly on urban spatial-temporal data due to ignoring spatial correlation. This suggests that incorporating spatial correlation is crucial to improve prediction accuracy.

(2) The spatial-CNN-based sequential structure models, such as STResNet, ACFM, and DMVSTNet, are only suitable for grid relation data as CNNs can not handle the graph relation data. Although these models are designed for grid data, they perform worse than the spatial-GCN-based models. One possible reason is that the original versions of such models require the input of information from more distant historical data, such as data from the same time period a day ago or a week ago. However, in our experiments, we have removed such information to ensure fair comparisons among the models. Another reason could be that graph neural networks have the ability to capture spatial correlations from grid data if an appropriate graph structure is constructed. This advantage could contribute to the better performance of spatial-GCN-based models. % in handling grid relation data.

(3) After analyzing the results from all the tables, we conclude that D2STGNN is the top-performing model. D2STGNN stands out due to its utilization of spatial-temporal synchronized learning and its systematic design of diffusion and inherent modules, tailored to the unique characteristics of spatial-temporal diffusion signals and inherent signals. In particular, each spatial-temporal signal consists of both a diffusion signal, which captures the information diffused from other sensors, and an inherent signal, which captures the information that is independent of other sensors. D2STGNN incorporates a residual decomposition mechanism that effectively separates the spatial-temporal signals. This allows more precise modeling of the different parts of spatial-temporal data to improve prediction accuracy.

% the separated signals to be processed independently by the diffusion and inherent modules, leading to exceptional performance.

(4) Moving on, MTGNN and GWNET are ranked 2nd and 3rd, respectively. Both models, MTGNN and GWNET, fall under the category of sequential structure models that combine Graph Convolutional Networks (GCNs) with multi-layer Dilated Casual Convolution (also known as WaveNet). They also incorporate an adaptive graph structure learning module within the graph convolution component to address the limitations of fixed graph structures used in DCRNN and STGCN. Additionally, these models employ direct prediction instead of a recurrent approach, which helps mitigate the issue of cumulative errors. MTGNN and GWNET consistently demonstrate strong performance across different datasets. This highlights the potential of combining GCNs with WaveNetS as a promising research direction for future advancements in macro spatial-temporal prediction. Several recent works have also adopted a similar structure, further reinforcing its effectiveness~\cite{DMSTGCN, ESG}.

(5) After this, the 4th ranked model is AGCRN. Both AGCRN and DCRNN employ the coupled structure, which combines Graph Convolutional Networks (GCNs) with Recurrent Neural Networks (RNNs). AGCRN outperforms DCRNN by introducing a learnable graph structure and adopting direct prediction instead of recurrent prediction. On the other hand, models that utilize sequential structures combining GCNs and RNNs, such as TGCN, achieve slightly lower performance. This suggests that the coupled structure is more suitable for integrating GCNs and RNNs compared to sequential structures. It is worth noting that the top 4 models in the overall ranking all use direct prediction. The direct prediction approach involves using two fully-connected layers on the learned hidden states to make multi-step predictions directly without relying on the results of previous steps for the recurrent prediction. This approach helps alleviate the problem of cumulative errors and is more suitable for short-term prediction scenarios such as the urban spatial-temporal prediction. 

(6) As a sequential structure model that combines spatial attention with other components, GMAN ranks 5th overall. GMAN utilizes temporal and spatial attention mechanisms to learn spatial-temporal correlations in the data, resulting in a good performance. However, the model has high complexity and consumes a large amount of memory. The other two attention-based models, namely STTN and ASTGCN, perform worse than GMAN, which could be attributed to differences in how attention is computed. GMAN employs a structure based entirely on attention mechanisms and introduces multi-head attention and group attention, while STTN and ASTGCN have a simpler implementation of attention. The main limitation of attention-based models is their high time complexity. Future studies can focus on reducing the complexity of the model and achieving a balance between performance and efficiency.

(7) Lastly, it is worth noting that other models employing spatial-temporal synchronized learning, such as STG2Seq and STSGCN, exhibit slightly lower performance. This can be attributed to their focus on short-range spatial-temporal correlations while overlooking long-range correlations. On the other hand, models like MTGNN and GWNET, which utilize the multi-layer WaveNet architecture, aim to capture long-range temporal correlations. Additionally, recent studies have also highlighted the significance of long-range spatial dependencies in spatial-temporal prediction~\cite{PDFormer, kddcup}. These findings further emphasize the importance of considering both short- and long-range correlations in spatial-temporal modeling and prediction tasks.

% Table generated by Excel2LaTeX from sheet 'Sheet2'
\begin{table*}[t]
\small
  \centering
  \caption{Comparison of Training and Inference Time Per Epoch (Unit: seconds)}
   \resizebox{\textwidth}{!}{
    \begin{tabular}{c|cc|cc|cc|cc|cc|cc|cc}
    \toprule
    Dataset & \multicolumn{2}{c|}{METR-LA} & \multicolumn{2}{c|}{PEMS-BAY} & \multicolumn{2}{c|}{PEMSD7(M)} & \multicolumn{2}{c|}{PEMSD3} & \multicolumn{2}{c|}{PEMSD4-Flow} & \multicolumn{2}{c|}{PEMSD8-Flow} & \multicolumn{2}{c}{RANK} \\
    \midrule
    Model & Train & Infer & Train & Infer & Train & Infer & Train & Infer & Train & Infer & Train & Infer & Train & Infer \\
    \midrule
    Seq2Seq & 20.94  & 0.86  & 32.85  & 1.35  & 7.14  & 0.33  & 13.32  & 0.69  & 8.41  & 0.43  & 10.16  & 0.48  & 3     & 3 \\
    AutoEncoder & 6.84  & 0.38  & 10.51  & 0.69  & 2.65  & 0.15  & 5.40  & 0.37  & 3.68  & 0.23  & 3.29  & 0.15  & 2     & 2 \\
    FNN   & 5.72  & 0.35  & 9.18  & 0.65  & 2.27  & 0.14  & 4.74  & 0.35  & 3.32  & 0.21  & 3.01  & 0.17  & 1     & 1 \\
    \midrule
    STGCN & 25.85  & 9.99  & 41.66  & 14.97  & 10.68  & 3.50  & 18.85  & 7.55  & 12.88  & 4.94  & 12.91  & 4.94  & 4     & 12 \\
    GWNET & 91.45  & 2.86  & 153.21  & 4.95  & 36.37  & 1.17  & 61.86  & 2.43  & 44.86  & 1.64  & 42.97  & 1.42  & 8     & 5 \\
    MTGNN & 75.64  & 2.73  & 121.95  & 4.29  & 27.14  & 0.99  & 60.15  & 2.29  & 38.22  & 1.45  & 37.67  & 1.28  & 5     & 4 \\
    TGCN  & 64.50  & 5.75  & 102.07  & 9.03  & 27.16  & 1.93  & 62.60  & 4.02  & 39.45  & 2.92  & 38.49  & 2.45  & 6     & 7 \\
    \midrule
    ASTGCN & 90.67  & 4.91  & 145.59  & 8.71  & 34.28  & 1.88  & 74.26  & 4.68  & 46.44  & 2.80  & 42.71  & 2.26  & 7     & 6 \\
    GMAN  & 738.24  & 14.57  & 1950.34  & 40.04  & 298.04  & 5.98  & 1116.30  & 22.57  & 579.96  & 11.88  & 326.05  & 6.34  & 15    & 14 \\
    STTN  & 166.38  & 7.51  & 274.66  & 14.62  & 61.09  & 2.83  & 145.01  & 7.83  & 86.71  & 4.55  & 84.52  & 3.66  & 11    & 9 \\
    \midrule
    DCRNN & 605.88  & 37.18  & 967.10  & 60.29  & 340.60  & 23.08  & 490.85  & 32.16  & 294.00  & 18.76  & 319.62  & 19.61  & 14    & 15 \\
    AGCRN & 156.36  & 8.24  & 251.72  & 15.41  & 59.22  & 3.12  & 128.18  & 7.16  & 82.38  & 4.92  & 82.39  & 4.08  & 9     & 10 \\
    \midrule
    STG2Seq & 164.39  & 9.24  & 261.81  & 15.52  & 59.87  & 3.27  & 135.30  & 7.92  & 85.77  & 4.78  & 84.23  & 4.31  & 10    & 11 \\
    STSGCN & 181.62  & 5.94  & 269.59  & 8.61  & 66.32  & 2.08  & 142.72  & 4.78  & 89.58  & 2.84  & 94.06  & 3.48  & 12    & 8 \\
    D2STGNN & 259.78  & 10.97  & 430.64  & 19.36  & 99.41  & 4.60  & 222.67  & 10.32  & 139.02  & 6.68  & 135.44  & 5.40  & 13    & 13 \\
    \midrule
    \midrule
    Dataset & \multicolumn{2}{c|}{NYCTaxi140103} & \multicolumn{2}{c|}{NYCTaxi160102} & \multicolumn{2}{c|}{NYCBike140409} & \multicolumn{2}{c|}{NYCBike160708} & \multicolumn{2}{c|}{TaxiBJ2014} & \multicolumn{2}{c|}{BikeCHI} & \multicolumn{2}{c}{RANK} \\
    \midrule
    Model & Train & Infer & Train & Infer & Train & Infer & Train & Infer & Train & Infer & Train & Infer & Train & Infer \\
    \midrule
    Seq2Seq & 0.40  & 0.02  & 0.73  & 0.04  & 0.97  & 0.04  & 0.62  & 0.03  & 1.25  & 0.09  & 0.48  & 0.03  & 3     & 3 \\
    AutoEncoder & 0.30  & 0.02  & 0.45  & 0.02  & 0.65  & 0.03  & 0.51  & 0.02  & 1.03  & 0.07  & 0.34  & 0.02  & 2     & 2 \\
    FNN   & 0.25  & 0.02  & 0.36  & 0.02  & 0.51  & 0.03  & 0.37  & 0.02  & 0.86  & 0.07  & 0.28  & 0.02  & 1     & 1 \\
    \midrule
    STResNet & 2.45  & 0.11  & 3.63  & 0.16  & 4.49  & 0.25  & 3.68  & 0.17  & 5.97  & 0.29  & 2.76  & 0.13  & 5     & 7 \\
    ACFM  & 14.23  & 0.52  & 15.69  & 0.68  & 23.17  & 1.05  & 18.67  & 0.65  & 25.40  & 1.13  & 12.02  & 0.53  & 16    & 15 \\
    DMVSTNet & 36.48  & 2.11  & 50.25  & 2.81  & 38.99  & 2.20  & 50.33  & 2.80  & 611.21  & 35.47  & 59.99  & 3.42  & 18    & 18 \\
    \midrule
    STGCN & 1.41  & 0.07  & 2.16  & 0.08  & 3.04  & 0.14  & 2.01  & 0.08  & 6.11  & 0.19  & 1.72  & 0.08  & 4     & 4 \\
    GWNET & 2.55  & 0.09  & 3.20  & 0.12  & 4.57  & 0.19  & 3.42  & 0.11  & 13.00  & 0.31  & 2.67  & 0.09  & 7     & 5 \\
    MTGNN & 2.71  & 0.10  & 3.19  & 0.13  & 5.38  & 0.18  & 3.61  & 0.13  & 8.83  & 0.57  & 2.55  & 0.09  & 8     & 6 \\
    TGCN  & 2.63  & 0.12  & 3.57  & 0.16  & 4.63  & 0.21  & 3.47  & 0.15  & 6.10  & 0.69  & 2.71  & 0.14  & 6     & 9 \\
    \midrule
    ASTGCN & 2.46  & 0.14  & 3.45  & 0.18  & 5.05  & 0.27  & 3.57  & 0.17  & 9.86  & 0.68  & 2.89  & 0.14  & 9     & 10 \\
    GMAN  & 21.00  & 0.65  & 28.12  & 0.84  & 36.72  & 1.28  & 28.67  & 0.88  & 172.37  & 6.06  & 26.18  & 0.69  & 17    & 17 \\
    STTN  & 6.77  & 0.28  & 9.57  & 0.37  & 14.53  & 0.57  & 9.43  & 0.38  & 47.61  & 2.56  & 7.49  & 0.33  & 14    & 14 \\
    \midrule
    DCRNN & 10.26  & 0.67  & 13.70  & 0.81  & 21.78  & 1.31  & 14.78  & 0.88  & 26.29  & 1.85  & 10.42  & 0.65  & 15    & 16 \\
    AGCRN & 2.79  & 0.13  & 3.13  & 0.18  & 5.63  & 0.28  & 3.72  & 0.18  & 11.21  & 0.37  & 2.45  & 0.15  & 10    & 11 \\
    \midrule
    STG2Seq & 3.23  & 0.14  & 3.88  & 0.16  & 5.92  & 0.23  & 3.82  & 0.15  & 10.92  & 0.69  & 3.07  & 0.13  & 11    & 8 \\
    STSGCN & 3.55  & 0.14  & 4.48  & 0.18  & 7.76  & 0.29  & 5.00  & 0.19  & 11.71  & 0.61  & 3.87  & 0.15  & 12    & 13 \\
    D2STGNN & 3.49  & 0.14  & 5.18  & 0.17  & 7.83  & 0.26  & 5.32  & 0.18  & 22.54  & 1.16  & 3.97  & 0.15  & 13    & 12 \\
    \bottomrule
    \end{tabular}%
    }
    \vspace{-0.2cm}
  \label{tab:effic}%
  \vspace{-0.3cm}
\end{table*}%

\renewcommand{\thesubfigure}{(\arabic{subfigure})}

\begin{figure*}[htbp]
    \centering
    \subfigure[METR-LA dataset]{
        \includegraphics[width=0.45\textwidth, page=1]{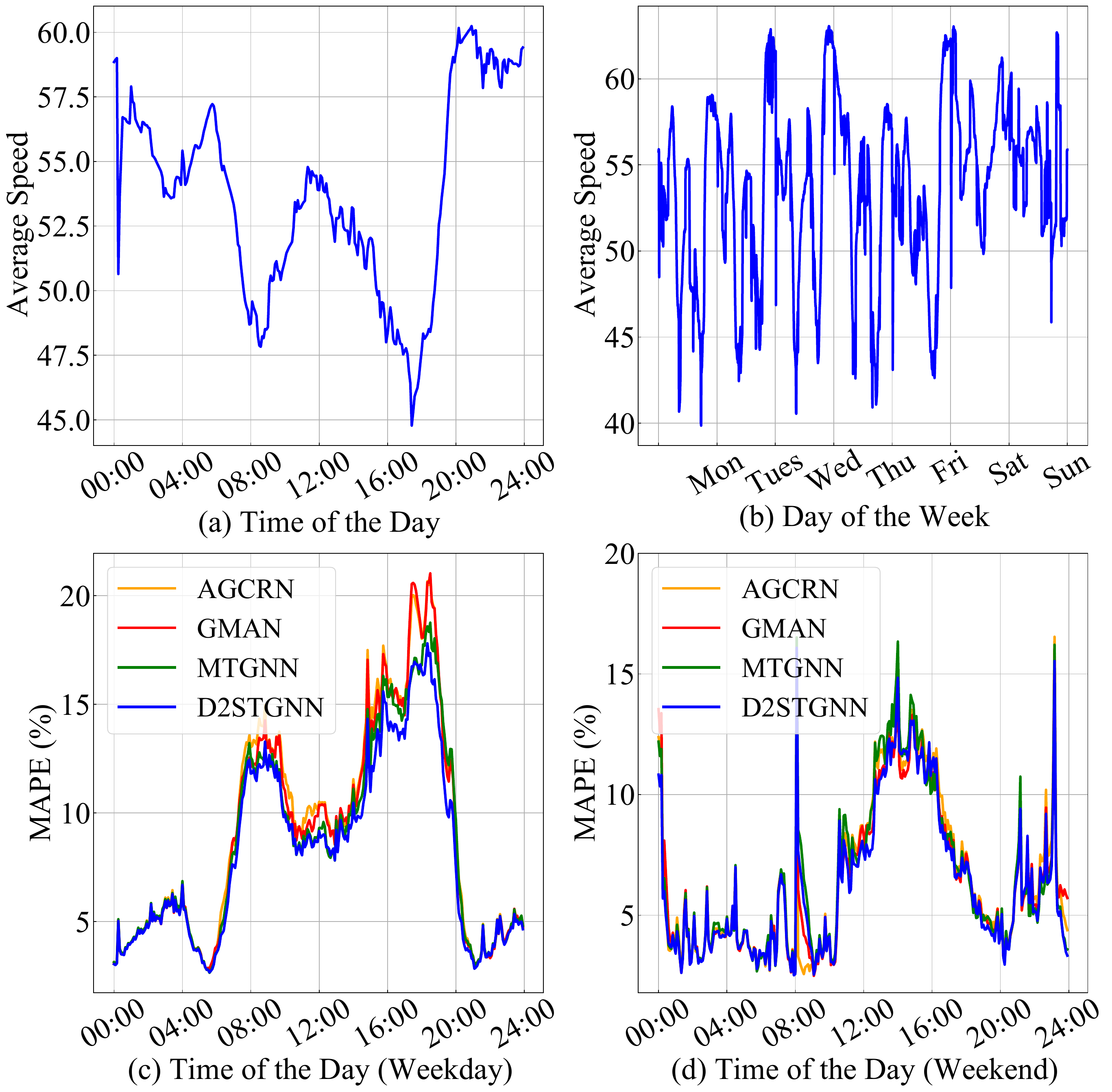}
        \label{fig:expa}
    }
    \subfigure[PEMSD8-Speed dataset]{
        \includegraphics[width=0.435\textwidth, page=2, page=1]{figures/PEMSD8_Speed.pdf}
        \label{fig:expb}
    }
    \vspace{-0.4cm}
    \caption{Data distribution and model performance of the METR-LA and PEMSD8-Speed at different time periods.}
    \label{fig:vis}
    \vspace{-0.5cm}
\end{figure*}

\vspace{-0.8cm}
\subsection{Model Efficiency Study}

Table~\ref{tab:effic} displays the training and inference time for each epoch, separately for graph relation data and grid relation data. Due to limited space, only the results of 12 datasets are provided here. However, it is important to note that other datasets typically exhibit similar efficiency performance to those shown in Table~\ref{tab:effic}, such as PEMSD4-Flow, PEMSD4-Occupy, and PEMSD4-Speed. Throughout the experiment, we use a consistent batch size of 16 to prevent GPU memory consumption from surpassing the capacity of a single card. Based on these results, we can make the following observations:

(1) General time series prediction models that do not consider spatial dependencies between spatial entities generally exhibit noticeably faster training and inference times compared to others.

(2) Among the three spatial-CNNs-based sequential structure models, STResNet stands out for its relatively high computational efficiency. This is primarily attributed to its utilization of convolution-based residual networks, which are computationally efficient compared to models incorporating RNNs and attention mechanisms like ACFM and DMVSTNet.

(3) In terms of other sequential structure models, spatial GCN-based models such as STGCN, GWNET, MTGNN and TGCN demonstrate superior computational efficiency in comparison to spatial-attention-based models such as GMAN and STTN, especially when dealing with large graphs such as TaxiBJ2014 (1024 nodes). This is primarily because spatial-attention-based models require calculating attention between each spatial entity and all other spatial entities, leading to higher computational costs. In contrast, spatial-GCN-based models typically focus solely on the spatial dependencies of neighboring spatial entities within a specific distance or number of hops for each spatial entity. As a result, they incur significantly lower computational costs. These spatial-attention-based models prioritize performance over computational efficiency. While they may have lower computational efficiency, they aim to deliver superior model performance.

(4) The sequential structure models such as STGCN, GWNET, and MTGNN tend to have shorter training and inference times compared to coupled structure models such as DCRNN and AGCRN in general. This is mainly due to the utilization of Temporal Convolutional Networks (TCNs) in sequential models, which generally offer faster computation compared to the Recurrent Neural Networks (RNNs) used in coupled structure models.

(5) When conducting multi-step predictions, recurrent prediction models such as DCRNN and GMAN exhibit significantly longer training and inference times compared to direct prediction models like GWNET, MTGNN, TGCN, and ASTGCN. The recurrent prediction models typically follow an encoder-decoder architecture and generate predictions based on previous predictions, making them unsuitable for parallel computation and resulting in increased computational complexity. However, the direct prediction models can predict the results of multiple steps at once, leading to faster training and inference times.

% However, the direct prediction models utilize two fully-connected layers or 1x1 convolutional layers on top of the learned hidden states of historical data, enabling efficient multi-step predictions in a single forward procedure. These models can benefit from fully parallel training by eliminating time-consuming recurrent structures, leading to faster training and inference times. It is worth noting that the direct prediction models not only excel in efficiency but also demonstrate superior performance. The top four models, in terms of performance, all employ the direct prediction methodologies.

(6) The relatively low efficiency of spatial-temporal synchronous modeling models, such as STG2Seq, STSGCN, and D2STGNN, can be attributed to the inclusion of more complex spatial-temporal joint learning modules. Only some of the larger models are less efficient than them, such as DCRNN and GMAN, but both are mainly limited by the paradigm of recurrent prediction. Although D2STGNN is the top model in terms of performance, its efficiency is far from that of MTGNN and GWNET, which rank 2nd and 3rd. When selecting a model for a particular application, it is critical to consider the requirements of the scenario, taking into account both efficiency and performance to make the appropriate decision.

\subsection{Analysis of Prediction Results and Performance Discrepancies}

We select the best-performing models from each category according to Table~\ref{tab:rank} and visualize their prediction results on the METR-LA and PEMSD8-Speed datasets for different times and days in Figure~\ref{fig:expa} and Figure~\ref{fig:expb} respectively. For each dataset, we plot four subplots: (a) and (b) show the trend of average traffic speed at different times of the day and different days of the week, respectively; (c) and (d) show the mean absolute percentage error (MAPE) between the model's prediction and the ground truth at different times on weekdays and weekends, respectively.

From subfigures (a) and (b), we observe peak periods and periodicity in the real data. The peak periods are typically concentrated around 8:00 and 17:00, while the data exhibits periodic patterns on weekdays and weekends, respectively. Furthermore, the traffic speed distribution on weekends appears more concentrated, with a smaller variance than on weekdays. Correspondingly, according to subfigures (c) and (d), there is a significant difference in the performance of the models on weekdays and weekends. The models perform better on weekends, which can be attributed to the more concentrated data distribution during those periods. On weekdays, however, the selected models achieve relatively similar performance during non-peak periods, while there is a noticeable performance gap during peak periods. The main source of the performance gap among these four models is their performance during peak periods.

As shown in subfigures (a) and (b), traffic conditions during peak periods tend to exhibit more complex and rapid changes, with a higher likelihood of extreme situations such as traffic congestion or accidents. Therefore, predicting traffic during peak periods is more challenging than during other periods. However, in practical applications, accurate traffic predictions during peak periods are precisely what people are more concerned about. These figures also indicate that improving the overall performance of the models relies on enhancing their prediction accuracy during peak periods. These observations provide researchers with a direction for the future design and development of better traffic models, focusing on capturing the fluctuations in traffic conditions during peak periods.

\section{Conclusion} \label{conclusion}
\balance
% In this work, we present a comprehensive review of urban spatial-temporal prediction and proposes a unified storage format for spatial-temporal data, called atomic files. Building on this, we introduce \name, a unified and comprehensive open-source library for urban spatial-temporal prediction that includes \rawDataCnt spatial-temporal datasets and \modelCnt spatial-temporal prediction models covering \taskcnt mainstream sub-tasks of urban spatial-temporal prediction. By conducting extensive experiments using \name, we establish a comprehensive model performance leaderboard that identifies promising research directions for spatial-temporal prediction.

% To the best of our knowledge, \name is the first open-source library for urban spatial-temporal prediction, providing a valuable tool for exploring and developing spatial-temporal prediction models. We will continuously expand \name to contribute to the spatial-temporal prediction field in the future. For example, we can cover more spatial-temporal prediction tasks, such as climate prediction, air quality prediction, theft prediction, etc.

\subsection{Limitations and Future Work}
It is important to acknowledge the limitations of this study. This paper primarily focuses on traffic prediction in the transportation field within the realm of urban spatial-temporal prediction while giving less attention to other related fields such as air quality prediction, climate prediction, and crime frequency prediction. This choice is driven by the fact that traffic prediction is one of the most critical and rapidly evolving domains within urban spatial-temporal prediction. However, it is essential to highlight that the proposed "atomic files" format is not exclusively limited to traffic data. It can be applied to various other types of urban spatial-temporal data, including temperature, wind speed, crime incidents, and more.

Exploring other domains of spatial-temporal prediction tasks, such as meteorology, is a potential avenue for future work. Expanding the scope of research to encompass a broader range of prediction tasks will contribute to a more comprehensive understanding of urban spatial-temporal dynamics and facilitate technique advancements in various fields. Future work will involve exploring additional areas of spatial-temporal prediction, expanding the utility of this work, and further enhancing our understanding of spatial-temporal dynamics in different contexts.

\subsection{Conclusion}
In conclusion, this work addresses the challenges faced in the field of urban spatial-temporal prediction and makes significant contributions to data management and prediction techniques. By introducing the unified storage format "atomic files," we simplify the access and utilization of diverse urban spatial-temporal datasets, enhancing data management efficiency. The comprehensive overview of technological advances in prediction models provides researchers with valuable insights and guidance for developing robust models. Additionally, the extensive experiments conducted using diverse models and datasets establish a performance benchmark and identify promising research directions.

Looking ahead, the findings and methodologies presented in this work pave the way for future research in urban spatial-temporal prediction. Researchers and practitioners can leverage the unified storage format and technological roadmap and model performance benchmark to explore new avenues, develop innovative models, and contribute to the advancement of the field. Ultimately, these efforts will lead to more effective management of urban spatial-temporal data, enhanced prediction capabilities, and improved urban planning and development.

\clearpage

\bibliographystyle{ACM-Reference-Format}
% \balance
\bibliography{sample}


\begin{thebibliography}{77}

%%% the Web address.

\ifx \showCODEN    \undefined \def \showCODEN     #1{\unskip}     \fi
\ifx \showDOI      \undefined \def \showDOI       #1{#1}\fi
\ifx \showISBNx    \undefined \def \showISBNx     #1{\unskip}     \fi
\ifx \showISBNxiii \undefined \def \showISBNxiii  #1{\unskip}     \fi
\ifx \showISSN     \undefined \def \showISSN      #1{\unskip}     \fi
\ifx \showLCCN     \undefined \def \showLCCN      #1{\unskip}     \fi
\ifx \shownote     \undefined \def \shownote      #1{#1}          \fi
\ifx \showarticletitle \undefined \def \showarticletitle #1{#1}   \fi
\ifx \showURL      \undefined \def \showURL       {\relax}        \fi
% invisible to TeX
\providecommand\bibfield[2]{#2}
\providecommand\bibinfo[2]{#2}
\providecommand\natexlab[1]{#1}
\providecommand\showeprint[2][]{arXiv:#2}

\bibitem[\protect\citeauthoryear{Bai, Yao, Kanhere, Wang, and Sheng}{Bai
  et~al\mbox{.}}{2019}]%
        {STG2Seq}
\bibfield{author}{\bibinfo{person}{Lei Bai}, \bibinfo{person}{Lina Yao},
  \bibinfo{person}{Salil~S. Kanhere}, \bibinfo{person}{Xianzhi Wang}, {and}
  \bibinfo{person}{Quan~Z. Sheng}.} \bibinfo{year}{2019}\natexlab{}.
\newblock \showarticletitle{STG2Seq: Spatial-Temporal Graph to Sequence Model
  for Multi-step Passenger Demand Forecasting}. In
  \bibinfo{booktitle}{\emph{{IJCAI}}}. \bibinfo{publisher}{ijcai.org},
  \bibinfo{pages}{1981--1987}.
\newblock


\bibitem[\protect\citeauthoryear{Bai, Yao, Li, Wang, and Wang}{Bai
  et~al\mbox{.}}{2020}]%
        {AGCRN}
\bibfield{author}{\bibinfo{person}{Lei Bai}, \bibinfo{person}{Lina Yao},
  \bibinfo{person}{Can Li}, \bibinfo{person}{Xianzhi Wang}, {and}
  \bibinfo{person}{Can Wang}.} \bibinfo{year}{2020}\natexlab{}.
\newblock \showarticletitle{Adaptive Graph Convolutional Recurrent Network for
  Traffic Forecasting}.
\newblock \bibinfo{journal}{\emph{NIPS}}  \bibinfo{volume}{33}
  (\bibinfo{year}{2020}).
\newblock


\bibitem[\protect\citeauthoryear{Chen, Chen, Xie, Cao, Gao, and Feng}{Chen
  et~al\mbox{.}}{2020}]%
        {MRABGCN}
\bibfield{author}{\bibinfo{person}{Weiqi Chen}, \bibinfo{person}{Ling Chen},
  \bibinfo{person}{Yu Xie}, \bibinfo{person}{Wei Cao}, \bibinfo{person}{Yusong
  Gao}, {and} \bibinfo{person}{Xiaojie Feng}.} \bibinfo{year}{2020}\natexlab{}.
\newblock \showarticletitle{Multi-Range Attentive Bicomponent Graph
  Convolutional Network for Traffic Forecasting}. In
  \bibinfo{booktitle}{\emph{{AAAI}}}. \bibinfo{publisher}{{AAAI} Press},
  \bibinfo{pages}{3529--3536}.
\newblock


\bibitem[\protect\citeauthoryear{Choi, Choi, Hwang, and Park}{Choi
  et~al\mbox{.}}{2022}]%
        {STG-NCDE}
\bibfield{author}{\bibinfo{person}{Jeongwhan Choi}, \bibinfo{person}{Hwangyong
  Choi}, \bibinfo{person}{Jeehyun Hwang}, {and} \bibinfo{person}{Noseong
  Park}.} \bibinfo{year}{2022}\natexlab{}.
\newblock \showarticletitle{Graph Neural Controlled Differential Equations for
  Traffic Forecasting}. In \bibinfo{booktitle}{\emph{{AAAI}}}.
  \bibinfo{publisher}{{AAAI} Press}, \bibinfo{pages}{6367--6374}.
\newblock


\bibitem[\protect\citeauthoryear{Chung, G{\"{u}}l{\c{c}}ehre, Cho, and
  Bengio}{Chung et~al\mbox{.}}{2014}]%
        {GRU}
\bibfield{author}{\bibinfo{person}{Junyoung Chung},
  \bibinfo{person}{{\c{C}}aglar G{\"{u}}l{\c{c}}ehre},
  \bibinfo{person}{KyungHyun Cho}, {and} \bibinfo{person}{Yoshua Bengio}.}
  \bibinfo{year}{2014}\natexlab{}.
\newblock \showarticletitle{Empirical Evaluation of Gated Recurrent Neural
  Networks on Sequence Modeling}.
\newblock \bibinfo{journal}{\emph{CoRR}}  \bibinfo{volume}{abs/1412.3555}
  (\bibinfo{year}{2014}).
\newblock


\bibitem[\protect\citeauthoryear{Cui, Zheng, Cui, Xie, Deng, Huang, and
  Zhou}{Cui et~al\mbox{.}}{2021}]%
        {METRO}
\bibfield{author}{\bibinfo{person}{Yue Cui}, \bibinfo{person}{Kai Zheng},
  \bibinfo{person}{Dingshan Cui}, \bibinfo{person}{Jiandong Xie},
  \bibinfo{person}{Liwei Deng}, \bibinfo{person}{Feiteng Huang}, {and}
  \bibinfo{person}{Xiaofang Zhou}.} \bibinfo{year}{2021}\natexlab{}.
\newblock \showarticletitle{{METRO:} {A} Generic Graph Neural Network Framework
  for Multivariate Time Series Forecasting}.
\newblock \bibinfo{journal}{\emph{Proc. {VLDB} Endow.}} \bibinfo{volume}{15},
  \bibinfo{number}{2} (\bibinfo{year}{2021}), \bibinfo{pages}{224--236}.
\newblock


\bibitem[\protect\citeauthoryear{Cui, Henrickson, Ke, and Wang}{Cui
  et~al\mbox{.}}{2019}]%
        {TGCLSTM}
\bibfield{author}{\bibinfo{person}{Zhiyong Cui}, \bibinfo{person}{Kristian
  Henrickson}, \bibinfo{person}{Ruimin Ke}, {and} \bibinfo{person}{Yinhai
  Wang}.} \bibinfo{year}{2019}\natexlab{}.
\newblock \showarticletitle{Traffic graph convolutional recurrent neural
  network: A deep learning framework for network-scale traffic learning and
  forecasting}.
\newblock \bibinfo{journal}{\emph{IEEE Transactions on Intelligent
  Transportation Systems}} \bibinfo{volume}{21}, \bibinfo{number}{11}
  (\bibinfo{year}{2019}), \bibinfo{pages}{4883--4894}.
\newblock


\bibitem[\protect\citeauthoryear{Cui, Ke, and Wang}{Cui et~al\mbox{.}}{2018}]%
        {cui2018deep}
\bibfield{author}{\bibinfo{person}{Zhiyong Cui}, \bibinfo{person}{Ruimin Ke},
  {and} \bibinfo{person}{Yinhai Wang}.} \bibinfo{year}{2018}\natexlab{}.
\newblock \showarticletitle{Deep Bidirectional and Unidirectional {LSTM}
  Recurrent Neural Network for Network-wide Traffic Speed Prediction}.
\newblock \bibinfo{journal}{\emph{CoRR}}  \bibinfo{volume}{abs/1801.02143}
  (\bibinfo{year}{2018}).
\newblock


\bibitem[\protect\citeauthoryear{de~Medrano and Aznarte}{de~Medrano and
  Aznarte}{2020}]%
        {de2020spatio}
\bibfield{author}{\bibinfo{person}{Rodrigo de Medrano} {and}
  \bibinfo{person}{Jos{\'{e}}~Luis Aznarte}.} \bibinfo{year}{2020}\natexlab{}.
\newblock \showarticletitle{A Spatio-Temporal Spot-Forecasting Framework
  forUrban Traffic Prediction}.
\newblock \bibinfo{journal}{\emph{CoRR}}  \bibinfo{volume}{abs/2003.13977}
  (\bibinfo{year}{2020}).
\newblock


\bibitem[\protect\citeauthoryear{Defferrard, Bresson, and
  Vandergheynst}{Defferrard et~al\mbox{.}}{2016}]%
        {ChebConv}
\bibfield{author}{\bibinfo{person}{Micha{\"{e}}l Defferrard},
  \bibinfo{person}{Xavier Bresson}, {and} \bibinfo{person}{Pierre
  Vandergheynst}.} \bibinfo{year}{2016}\natexlab{}.
\newblock \showarticletitle{Convolutional Neural Networks on Graphs with Fast
  Localized Spectral Filtering}. In \bibinfo{booktitle}{\emph{{NIPS}}}.
  \bibinfo{pages}{3837--3845}.
\newblock


\bibitem[\protect\citeauthoryear{Diao, Wang, Zhang, Liu, Xie, and He}{Diao
  et~al\mbox{.}}{2019}]%
        {DGCNN}
\bibfield{author}{\bibinfo{person}{Zulong Diao}, \bibinfo{person}{Xin Wang},
  \bibinfo{person}{Dafang Zhang}, \bibinfo{person}{Yingru Liu},
  \bibinfo{person}{Kun Xie}, {and} \bibinfo{person}{Shaoyao He}.}
  \bibinfo{year}{2019}\natexlab{}.
\newblock \showarticletitle{Dynamic Spatial-Temporal Graph Convolutional Neural
  Networks for Traffic Forecasting}. In \bibinfo{booktitle}{\emph{{AAAI}}}.
  \bibinfo{publisher}{{AAAI} Press}, \bibinfo{pages}{890--897}.
\newblock


\bibitem[\protect\citeauthoryear{Drucker, Burges, Kaufman, Smola, and
  Vapnik}{Drucker et~al\mbox{.}}{1996}]%
        {SVR}
\bibfield{author}{\bibinfo{person}{Harris Drucker},
  \bibinfo{person}{Christopher J.~C. Burges}, \bibinfo{person}{Linda Kaufman},
  \bibinfo{person}{Alexander~J. Smola}, {and} \bibinfo{person}{Vladimir
  Vapnik}.} \bibinfo{year}{1996}\natexlab{}.
\newblock \showarticletitle{Support Vector Regression Machines}. In
  \bibinfo{booktitle}{\emph{{NIPS}}}. \bibinfo{publisher}{{MIT} Press},
  \bibinfo{pages}{155--161}.
\newblock


\bibitem[\protect\citeauthoryear{Fang, Long, Song, and Xie}{Fang
  et~al\mbox{.}}{2021a}]%
        {STGODE}
\bibfield{author}{\bibinfo{person}{Zheng Fang}, \bibinfo{person}{Qingqing
  Long}, \bibinfo{person}{Guojie Song}, {and} \bibinfo{person}{Kunqing Xie}.}
  \bibinfo{year}{2021}\natexlab{a}.
\newblock \showarticletitle{Spatial-Temporal Graph {ODE} Networks for Traffic
  Flow Forecasting}. In \bibinfo{booktitle}{\emph{{KDD}}}.
  \bibinfo{publisher}{{ACM}}, \bibinfo{pages}{364--373}.
\newblock


\bibitem[\protect\citeauthoryear{Fang, Pan, Chen, Du, and Gao}{Fang
  et~al\mbox{.}}{2021b}]%
        {MDTP}
\bibfield{author}{\bibinfo{person}{Ziquan Fang}, \bibinfo{person}{Lu Pan},
  \bibinfo{person}{Lu Chen}, \bibinfo{person}{Yuntao Du}, {and}
  \bibinfo{person}{Yunjun Gao}.} \bibinfo{year}{2021}\natexlab{b}.
\newblock \showarticletitle{{MDTP:} {A} Multi-source Deep Traffic Prediction
  Framework over Spatio-Temporal Trajectory Data}.
\newblock \bibinfo{journal}{\emph{Proc. {VLDB} Endow.}} \bibinfo{volume}{14},
  \bibinfo{number}{8} (\bibinfo{year}{2021}), \bibinfo{pages}{1289--1297}.
\newblock


\bibitem[\protect\citeauthoryear{Gu, Hu, and Jia}{Gu et~al\mbox{.}}{2023}]%
        {control}
\bibfield{author}{\bibinfo{person}{Ke Gu}, \bibinfo{person}{Jieyu Hu}, {and}
  \bibinfo{person}{Weijia Jia}.} \bibinfo{year}{2023}\natexlab{}.
\newblock \showarticletitle{Adaptive Area-Based Traffic Congestion Control and
  Management Scheme Based on Fog Computing}.
\newblock \bibinfo{journal}{\emph{{IEEE} Trans. Intell. Transp. Syst.}}
  \bibinfo{volume}{24}, \bibinfo{number}{1} (\bibinfo{year}{2023}),
  \bibinfo{pages}{1359--1373}.
\newblock


\bibitem[\protect\citeauthoryear{Guo, Hu, Qian, Sun, Gao, and Yin}{Guo
  et~al\mbox{.}}{2022a}]%
        {DGCN}
\bibfield{author}{\bibinfo{person}{Kan Guo}, \bibinfo{person}{Yongli Hu},
  \bibinfo{person}{Zhen~Sean Qian}, \bibinfo{person}{Yanfeng Sun},
  \bibinfo{person}{Junbin Gao}, {and} \bibinfo{person}{Baocai Yin}.}
  \bibinfo{year}{2022}\natexlab{a}.
\newblock \showarticletitle{Dynamic Graph Convolution Network for Traffic
  Forecasting Based on Latent Network of Laplace Matrix Estimation}.
\newblock \bibinfo{journal}{\emph{{IEEE} Trans. Intell. Transp. Syst.}}
  \bibinfo{volume}{23}, \bibinfo{number}{2} (\bibinfo{year}{2022}),
  \bibinfo{pages}{1009--1018}.
\newblock


\bibitem[\protect\citeauthoryear{Guo, Hu, Sun, Qian, Gao, and Yin}{Guo
  et~al\mbox{.}}{2021}]%
        {HGCN}
\bibfield{author}{\bibinfo{person}{Kan Guo}, \bibinfo{person}{Yongli Hu},
  \bibinfo{person}{Yanfeng Sun}, \bibinfo{person}{Sean Qian},
  \bibinfo{person}{Junbin Gao}, {and} \bibinfo{person}{Baocai Yin}.}
  \bibinfo{year}{2021}\natexlab{}.
\newblock \showarticletitle{Hierarchical graph convolution network for traffic
  forecasting}. In \bibinfo{booktitle}{\emph{Proceedings of the AAAI conference
  on artificial intelligence}}, Vol.~\bibinfo{volume}{35}.
  \bibinfo{pages}{151--159}.
\newblock


\bibitem[\protect\citeauthoryear{Guo, Lin, Feng, Song, and Wan}{Guo
  et~al\mbox{.}}{2019}]%
        {ASTGCN}
\bibfield{author}{\bibinfo{person}{Shengnan Guo}, \bibinfo{person}{Youfang
  Lin}, \bibinfo{person}{Ning Feng}, \bibinfo{person}{Chao Song}, {and}
  \bibinfo{person}{Huaiyu Wan}.} \bibinfo{year}{2019}\natexlab{}.
\newblock \showarticletitle{Attention based spatial-temporal graph
  convolutional networks for traffic flow forecasting}. In
  \bibinfo{booktitle}{\emph{Proceedings of the AAAI Conference on Artificial
  Intelligence}}, Vol.~\bibinfo{volume}{33}. \bibinfo{pages}{922--929}.
\newblock


\bibitem[\protect\citeauthoryear{Guo, Lin, Wan, Li, and Cong}{Guo
  et~al\mbox{.}}{2022b}]%
        {ASTGNN}
\bibfield{author}{\bibinfo{person}{Shengnan Guo}, \bibinfo{person}{Youfang
  Lin}, \bibinfo{person}{Huaiyu Wan}, \bibinfo{person}{Xiucheng Li}, {and}
  \bibinfo{person}{Gao Cong}.} \bibinfo{year}{2022}\natexlab{b}.
\newblock \showarticletitle{Learning Dynamics and Heterogeneity of
  Spatial-Temporal Graph Data for Traffic Forecasting}.
\newblock \bibinfo{journal}{\emph{{IEEE} Trans. Knowl. Data Eng.}}
  \bibinfo{volume}{34}, \bibinfo{number}{11} (\bibinfo{year}{2022}),
  \bibinfo{pages}{5415--5428}.
\newblock


\bibitem[\protect\citeauthoryear{Guopeng, Knoop, and van Lint}{Guopeng
  et~al\mbox{.}}{2020}]%
        {guopeng2020dynamic}
\bibfield{author}{\bibinfo{person}{LI Guopeng}, \bibinfo{person}{Victor~L
  Knoop}, {and} \bibinfo{person}{Hans van Lint}.}
  \bibinfo{year}{2020}\natexlab{}.
\newblock \showarticletitle{Dynamic Graph Filters Networks: A Gray-box Model
  for Multistep Traffic Forecasting}. In \bibinfo{booktitle}{\emph{2020 IEEE
  23rd International Conference on Intelligent Transportation Systems (ITSC)}}.
  IEEE, \bibinfo{pages}{1--6}.
\newblock


\bibitem[\protect\citeauthoryear{Hamilton}{Hamilton}{1994}]%
        {VAR}
\bibfield{author}{\bibinfo{person}{James~Douglas Hamilton}.}
  \bibinfo{year}{1994}\natexlab{}.
\newblock \bibinfo{booktitle}{\emph{Time series analysis}}.
\newblock \bibinfo{publisher}{Princeton university press}.
\newblock


\bibitem[\protect\citeauthoryear{Han, Du, Sun, Fu, Lv, and Xiong}{Han
  et~al\mbox{.}}{2021}]%
        {DMSTGCN}
\bibfield{author}{\bibinfo{person}{Liangzhe Han}, \bibinfo{person}{Bowen Du},
  \bibinfo{person}{Leilei Sun}, \bibinfo{person}{Yanjie Fu},
  \bibinfo{person}{Yisheng Lv}, {and} \bibinfo{person}{Hui Xiong}.}
  \bibinfo{year}{2021}\natexlab{}.
\newblock \showarticletitle{Dynamic and Multi-faceted Spatio-temporal Deep
  Learning for Traffic Speed Forecasting}. In
  \bibinfo{booktitle}{\emph{{KDD}}}. \bibinfo{publisher}{{ACM}},
  \bibinfo{pages}{547--555}.
\newblock


\bibitem[\protect\citeauthoryear{Han, Ma, Sun, Du, Fu, Lv, and Xiong}{Han
  et~al\mbox{.}}{2022}]%
        {CMOD}
\bibfield{author}{\bibinfo{person}{Liangzhe Han}, \bibinfo{person}{Xiaojian
  Ma}, \bibinfo{person}{Leilei Sun}, \bibinfo{person}{Bowen Du},
  \bibinfo{person}{Yanjie Fu}, \bibinfo{person}{Weifeng Lv}, {and}
  \bibinfo{person}{Hui Xiong}.} \bibinfo{year}{2022}\natexlab{}.
\newblock \showarticletitle{Continuous-Time and Multi-Level Graph
  Representation Learning for Origin-Destination Demand Prediction}. In
  \bibinfo{booktitle}{\emph{{KDD}}}. \bibinfo{publisher}{{ACM}},
  \bibinfo{pages}{516--524}.
\newblock


\bibitem[\protect\citeauthoryear{He, Li, Tan, Wu, and Li}{He
  et~al\mbox{.}}{2023}]%
        {OneShotSTL}
\bibfield{author}{\bibinfo{person}{Xiao He}, \bibinfo{person}{Ye Li},
  \bibinfo{person}{Jian Tan}, \bibinfo{person}{Bin Wu}, {and}
  \bibinfo{person}{Feifei Li}.} \bibinfo{year}{2023}\natexlab{}.
\newblock \showarticletitle{OneShotSTL: One-Shot Seasonal-Trend Decomposition
  For Online Time Series Anomaly Detection And Forecasting}.
\newblock \bibinfo{journal}{\emph{Proc. {VLDB} Endow.}} \bibinfo{volume}{16},
  \bibinfo{number}{6} (\bibinfo{year}{2023}), \bibinfo{pages}{1399--1412}.
\newblock


\bibitem[\protect\citeauthoryear{Hochreiter and Schmidhuber}{Hochreiter and
  Schmidhuber}{1997}]%
        {LSTM}
\bibfield{author}{\bibinfo{person}{Sepp Hochreiter} {and}
  \bibinfo{person}{J{\"{u}}rgen Schmidhuber}.} \bibinfo{year}{1997}\natexlab{}.
\newblock \showarticletitle{Long Short-Term Memory}.
\newblock \bibinfo{journal}{\emph{Neural Comput.}} \bibinfo{volume}{9},
  \bibinfo{number}{8} (\bibinfo{year}{1997}), \bibinfo{pages}{1735--1780}.
\newblock


\bibitem[\protect\citeauthoryear{Ji, Wang, Jiang, Jiang, and Zhang}{Ji
  et~al\mbox{.}}{2022}]%
        {STDEN}
\bibfield{author}{\bibinfo{person}{Jiahao Ji}, \bibinfo{person}{Jingyuan Wang},
  \bibinfo{person}{Zhe Jiang}, \bibinfo{person}{Jiawei Jiang}, {and}
  \bibinfo{person}{Hu Zhang}.} \bibinfo{year}{2022}\natexlab{}.
\newblock \showarticletitle{{STDEN:} Towards Physics-Guided Neural Networks for
  Traffic Flow Prediction}. In \bibinfo{booktitle}{\emph{{AAAI}}}.
  \bibinfo{publisher}{{AAAI} Press}, \bibinfo{pages}{4048--4056}.
\newblock


\bibitem[\protect\citeauthoryear{Ji, Wang, Jiang, Ma, and Zhang}{Ji
  et~al\mbox{.}}{2020}]%
        {STPEF}
\bibfield{author}{\bibinfo{person}{Jiahao Ji}, \bibinfo{person}{Jingyuan Wang},
  \bibinfo{person}{Zhe Jiang}, \bibinfo{person}{Jingtian Ma}, {and}
  \bibinfo{person}{Hu Zhang}.} \bibinfo{year}{2020}\natexlab{}.
\newblock \showarticletitle{Interpretable Spatiotemporal Deep Learning Model
  for Traffic Flow Prediction based on Potential Energy Fields}. In
  \bibinfo{booktitle}{\emph{{ICDM}}}. \bibinfo{publisher}{{IEEE}},
  \bibinfo{pages}{1076--1081}.
\newblock


\bibitem[\protect\citeauthoryear{Jiang, Han, and Wang}{Jiang
  et~al\mbox{.}}{2023a}]%
        {kddcup}
\bibfield{author}{\bibinfo{person}{Jiawei Jiang}, \bibinfo{person}{Chengkai
  Han}, {and} \bibinfo{person}{Jingyuan Wang}.}
  \bibinfo{year}{2023}\natexlab{a}.
\newblock \showarticletitle{BUAA\_BIGSCity: Spatial-Temporal Graph Neural
  Network for Wind Power Forecasting in Baidu KDD CUP 2022}.
\newblock \bibinfo{journal}{\emph{arXiv preprint arXiv:2302.11159}}
  (\bibinfo{year}{2023}).
\newblock


\bibitem[\protect\citeauthoryear{Jiang, Han, Zhao, and Wang}{Jiang
  et~al\mbox{.}}{2023b}]%
        {PDFormer}
\bibfield{author}{\bibinfo{person}{Jiawei Jiang}, \bibinfo{person}{Chengkai
  Han}, \bibinfo{person}{Wayne~Xin Zhao}, {and} \bibinfo{person}{Jingyuan
  Wang}.} \bibinfo{year}{2023}\natexlab{b}.
\newblock \showarticletitle{PDFormer: Propagation Delay-aware Dynamic
  Long-range Transformer for Traffic Flow Prediction}. In
  \bibinfo{booktitle}{\emph{{AAAI}}}. \bibinfo{publisher}{{AAAI} Press}.
\newblock


\bibitem[\protect\citeauthoryear{Li, Dai, Chen, Song, Zang, Huang, Lin, and
  Cai}{Li et~al\mbox{.}}{2023}]%
        {routeplanning}
\bibfield{author}{\bibinfo{person}{Bohan Li}, \bibinfo{person}{Tianlun Dai},
  \bibinfo{person}{Weitong Chen}, \bibinfo{person}{Xinyang Song},
  \bibinfo{person}{Yalei Zang}, \bibinfo{person}{Zhelong Huang},
  \bibinfo{person}{Qinyong Lin}, {and} \bibinfo{person}{Ken Cai}.}
  \bibinfo{year}{2023}\natexlab{}.
\newblock \showarticletitle{{T-PORP:} {A} Trusted Parallel Route Planning Model
  on Dynamic Road Networks}.
\newblock \bibinfo{journal}{\emph{{IEEE} Trans. Intell. Transp. Syst.}}
  \bibinfo{volume}{24}, \bibinfo{number}{1} (\bibinfo{year}{2023}),
  \bibinfo{pages}{1238--1250}.
\newblock


\bibitem[\protect\citeauthoryear{Li and Zhu}{Li and Zhu}{2021}]%
        {STFGNN}
\bibfield{author}{\bibinfo{person}{Mengzhang Li} {and}
  \bibinfo{person}{Zhanxing Zhu}.} \bibinfo{year}{2021}\natexlab{}.
\newblock \showarticletitle{Spatial-Temporal Fusion Graph Neural Networks for
  Traffic Flow Forecasting}. In \bibinfo{booktitle}{\emph{{AAAI}}}.
  \bibinfo{publisher}{{AAAI} Press}, \bibinfo{pages}{4189--4196}.
\newblock


\bibitem[\protect\citeauthoryear{Li, Yu, Shahabi, and Liu}{Li
  et~al\mbox{.}}{2018}]%
        {DCRNN}
\bibfield{author}{\bibinfo{person}{Yaguang Li}, \bibinfo{person}{Rose Yu},
  \bibinfo{person}{Cyrus Shahabi}, {and} \bibinfo{person}{Yan Liu}.}
  \bibinfo{year}{2018}\natexlab{}.
\newblock \showarticletitle{Diffusion Convolutional Recurrent Neural Network:
  Data-Driven Traffic Forecasting}. In \bibinfo{booktitle}{\emph{International
  Conference on Learning Representations (ICLR '18)}}.
\newblock


\bibitem[\protect\citeauthoryear{Liao, Zhang, Wu, McIlwraith, Chen, Yang, Guo,
  and Wu}{Liao et~al\mbox{.}}{2018}]%
        {bbliaojqZhangKDD18deep}
\bibfield{author}{\bibinfo{person}{Binbing Liao}, \bibinfo{person}{Jingqing
  Zhang}, \bibinfo{person}{Chao Wu}, \bibinfo{person}{Douglas McIlwraith},
  \bibinfo{person}{Tong Chen}, \bibinfo{person}{Shengwen Yang},
  \bibinfo{person}{Yike Guo}, {and} \bibinfo{person}{Fei Wu}.}
  \bibinfo{year}{2018}\natexlab{}.
\newblock \showarticletitle{Deep Sequence Learning with Auxiliary Information
  for Traffic Prediction}. In \bibinfo{booktitle}{\emph{Proceedings of the 24th
  ACM SIGKDD International Conference on Knowledge Discovery and Data Mining}}.
  ACM, \bibinfo{pages}{537--546}.
\newblock


\bibitem[\protect\citeauthoryear{Lin, Bai, Jia, Yang, and You}{Lin
  et~al\mbox{.}}{2020}]%
        {DSAN}
\bibfield{author}{\bibinfo{person}{Haoxing Lin}, \bibinfo{person}{Rufan Bai},
  \bibinfo{person}{Weijia Jia}, \bibinfo{person}{Xinyu Yang}, {and}
  \bibinfo{person}{Yongjian You}.} \bibinfo{year}{2020}\natexlab{}.
\newblock \showarticletitle{Preserving Dynamic Attention for Long-Term
  Spatial-Temporal Prediction}. In \bibinfo{booktitle}{\emph{{KDD}}}.
  \bibinfo{publisher}{{ACM}}, \bibinfo{pages}{36--46}.
\newblock


\bibitem[\protect\citeauthoryear{Liu, Wang, Shang, and Han}{Liu
  et~al\mbox{.}}{2022}]%
        {MSDR}
\bibfield{author}{\bibinfo{person}{Dachuan Liu}, \bibinfo{person}{Jin Wang},
  \bibinfo{person}{Shuo Shang}, {and} \bibinfo{person}{Peng Han}.}
  \bibinfo{year}{2022}\natexlab{}.
\newblock \showarticletitle{{MSDR:} Multi-Step Dependency Relation Networks for
  Spatial Temporal Forecasting}. In \bibinfo{booktitle}{\emph{{KDD}}}.
  \bibinfo{publisher}{{ACM}}, \bibinfo{pages}{1042--1050}.
\newblock


\bibitem[\protect\citeauthoryear{Liu, Chen, Wu, Zhen, Li, and Lin}{Liu
  et~al\mbox{.}}{2020}]%
        {liu2020physical}
\bibfield{author}{\bibinfo{person}{Lingbo Liu}, \bibinfo{person}{Jingwen Chen},
  \bibinfo{person}{Hefeng Wu}, \bibinfo{person}{Jiajie Zhen},
  \bibinfo{person}{Guanbin Li}, {and} \bibinfo{person}{Liang Lin}.}
  \bibinfo{year}{2020}\natexlab{}.
\newblock \showarticletitle{Physical-Virtual Collaboration Modeling for
  Intra-and Inter-Station Metro Ridership Prediction}.
\newblock \bibinfo{journal}{\emph{IEEE Transactions on Intelligent
  Transportation Systems}} (\bibinfo{year}{2020}).
\newblock


\bibitem[\protect\citeauthoryear{Liu, Qiu, Li, Wang, Ouyang, and Lin}{Liu
  et~al\mbox{.}}{2019}]%
        {CSTN}
\bibfield{author}{\bibinfo{person}{Lingbo Liu}, \bibinfo{person}{Zhilin Qiu},
  \bibinfo{person}{Guanbin Li}, \bibinfo{person}{Qing Wang},
  \bibinfo{person}{Wanli Ouyang}, {and} \bibinfo{person}{Liang Lin}.}
  \bibinfo{year}{2019}\natexlab{}.
\newblock \showarticletitle{Contextualized spatial--temporal network for taxi
  origin-destination demand prediction}.
\newblock \bibinfo{journal}{\emph{IEEE Transactions on Intelligent
  Transportation Systems}} \bibinfo{volume}{20}, \bibinfo{number}{10}
  (\bibinfo{year}{2019}), \bibinfo{pages}{3875--3887}.
\newblock


\bibitem[\protect\citeauthoryear{Liu, Zhang, Peng, Li, Du, and Lin}{Liu
  et~al\mbox{.}}{2018}]%
        {ACFM}
\bibfield{author}{\bibinfo{person}{Lingbo Liu}, \bibinfo{person}{Ruimao Zhang},
  \bibinfo{person}{Jiefeng Peng}, \bibinfo{person}{Guanbin Li},
  \bibinfo{person}{Bowen Du}, {and} \bibinfo{person}{Liang Lin}.}
  \bibinfo{year}{2018}\natexlab{}.
\newblock \showarticletitle{Attentive crowd flow machines}. In
  \bibinfo{booktitle}{\emph{Proceedings of the 26th ACM international
  conference on Multimedia}}. \bibinfo{pages}{1553--1561}.
\newblock


\bibitem[\protect\citeauthoryear{Lu, Gan, Jin, Fu, and Zhang}{Lu
  et~al\mbox{.}}{2020}]%
        {STAGGCN}
\bibfield{author}{\bibinfo{person}{Bin Lu}, \bibinfo{person}{Xiaoying Gan},
  \bibinfo{person}{Haiming Jin}, \bibinfo{person}{Luoyi Fu}, {and}
  \bibinfo{person}{Haisong Zhang}.} \bibinfo{year}{2020}\natexlab{}.
\newblock \showarticletitle{Spatiotemporal Adaptive Gated Graph Convolution
  Network for Urban Traffic Flow Forecasting}. In
  \bibinfo{booktitle}{\emph{{CIKM}}}. \bibinfo{publisher}{{ACM}},
  \bibinfo{pages}{1025--1034}.
\newblock


\bibitem[\protect\citeauthoryear{Lv, Duan, Kang, Li, and Wang}{Lv
  et~al\mbox{.}}{2014}]%
        {AutoEncoder}
\bibfield{author}{\bibinfo{person}{Yisheng Lv}, \bibinfo{person}{Yanjie Duan},
  \bibinfo{person}{Wenwen Kang}, \bibinfo{person}{Zhengxi Li}, {and}
  \bibinfo{person}{Fei-Yue Wang}.} \bibinfo{year}{2014}\natexlab{}.
\newblock \showarticletitle{Traffic flow prediction with big data: a deep
  learning approach}.
\newblock \bibinfo{journal}{\emph{IEEE Transactions on Intelligent
  Transportation Systems}} \bibinfo{volume}{16}, \bibinfo{number}{2}
  (\bibinfo{year}{2014}), \bibinfo{pages}{865--873}.
\newblock


\bibitem[\protect\citeauthoryear{Oreshkin, Amini, Coyle, and Coates}{Oreshkin
  et~al\mbox{.}}{2021}]%
        {FCGAGA}
\bibfield{author}{\bibinfo{person}{Boris~N. Oreshkin}, \bibinfo{person}{Arezou
  Amini}, \bibinfo{person}{Lucy Coyle}, {and} \bibinfo{person}{Mark Coates}.}
  \bibinfo{year}{2021}\natexlab{}.
\newblock \showarticletitle{{FC-GAGA:} Fully Connected Gated Graph Architecture
  for Spatio-Temporal Traffic Forecasting}. In
  \bibinfo{booktitle}{\emph{{AAAI}}}. \bibinfo{publisher}{{AAAI} Press},
  \bibinfo{pages}{9233--9241}.
\newblock


\bibitem[\protect\citeauthoryear{Paszke, Gross, Massa, Lerer, Bradbury, Chanan,
  Killeen, Lin, Gimelshein, Antiga, Desmaison, K{\"{o}}pf, Yang, DeVito,
  Raison, Tejani, Chilamkurthy, Steiner, Fang, Bai, and Chintala}{Paszke
  et~al\mbox{.}}{2019}]%
        {pytorch}
\bibfield{author}{\bibinfo{person}{Adam Paszke}, \bibinfo{person}{Sam Gross},
  \bibinfo{person}{Francisco Massa}, \bibinfo{person}{Adam Lerer},
  \bibinfo{person}{James Bradbury}, \bibinfo{person}{Gregory Chanan},
  \bibinfo{person}{Trevor Killeen}, \bibinfo{person}{Zeming Lin},
  \bibinfo{person}{Natalia Gimelshein}, \bibinfo{person}{Luca Antiga},
  \bibinfo{person}{Alban Desmaison}, \bibinfo{person}{Andreas K{\"{o}}pf},
  \bibinfo{person}{Edward~Z. Yang}, \bibinfo{person}{Zachary DeVito},
  \bibinfo{person}{Martin Raison}, \bibinfo{person}{Alykhan Tejani},
  \bibinfo{person}{Sasank Chilamkurthy}, \bibinfo{person}{Benoit Steiner},
  \bibinfo{person}{Lu Fang}, \bibinfo{person}{Junjie Bai}, {and}
  \bibinfo{person}{Soumith Chintala}.} \bibinfo{year}{2019}\natexlab{}.
\newblock \showarticletitle{PyTorch: An Imperative Style, High-Performance Deep
  Learning Library}. In \bibinfo{booktitle}{\emph{NeurIPS}}.
  \bibinfo{pages}{8024--8035}.
\newblock


\bibitem[\protect\citeauthoryear{Qiu, Zheng, Msahli, Memmi, Qiu, and Lu}{Qiu
  et~al\mbox{.}}{2021}]%
        {ToGCN}
\bibfield{author}{\bibinfo{person}{Han Qiu}, \bibinfo{person}{Qinkai Zheng},
  \bibinfo{person}{Mounira Msahli}, \bibinfo{person}{G{\'{e}}rard Memmi},
  \bibinfo{person}{Meikang Qiu}, {and} \bibinfo{person}{Jialiang Lu}.}
  \bibinfo{year}{2021}\natexlab{}.
\newblock \showarticletitle{Topological Graph Convolutional Network-Based Urban
  Traffic Flow and Density Prediction}.
\newblock \bibinfo{journal}{\emph{{IEEE} Trans. Intell. Transp. Syst.}}
  \bibinfo{volume}{22}, \bibinfo{number}{7} (\bibinfo{year}{2021}),
  \bibinfo{pages}{4560--4569}.
\newblock


\bibitem[\protect\citeauthoryear{Shao, Zhang, Wei, Wang, Xu, Cao, and
  Jensen}{Shao et~al\mbox{.}}{2022}]%
        {D2STGNN}
\bibfield{author}{\bibinfo{person}{Zezhi Shao}, \bibinfo{person}{Zhao Zhang},
  \bibinfo{person}{Wei Wei}, \bibinfo{person}{Fei Wang},
  \bibinfo{person}{Yongjun Xu}, \bibinfo{person}{Xin Cao}, {and}
  \bibinfo{person}{Christian~S. Jensen}.} \bibinfo{year}{2022}\natexlab{}.
\newblock \showarticletitle{Decoupled Dynamic Spatial-Temporal Graph Neural
  Network for Traffic Forecasting}.
\newblock \bibinfo{journal}{\emph{Proc. {VLDB} Endow.}} \bibinfo{volume}{15},
  \bibinfo{number}{11} (\bibinfo{year}{2022}), \bibinfo{pages}{2733--2746}.
\newblock


\bibitem[\protect\citeauthoryear{Song, Lin, Guo, and Wan}{Song
  et~al\mbox{.}}{2020}]%
        {STSGCN}
\bibfield{author}{\bibinfo{person}{Chao Song}, \bibinfo{person}{Youfang Lin},
  \bibinfo{person}{Shengnan Guo}, {and} \bibinfo{person}{Huaiyu Wan}.}
  \bibinfo{year}{2020}\natexlab{}.
\newblock \showarticletitle{Spatial-temporal synchronous graph convolutional
  networks: A new framework for spatial-temporal network data forecasting}. In
  \bibinfo{booktitle}{\emph{Proceedings of the AAAI Conference on Artificial
  Intelligence}}, Vol.~\bibinfo{volume}{34}. \bibinfo{pages}{914--921}.
\newblock


\bibitem[\protect\citeauthoryear{Sutskever, Vinyals, and Le}{Sutskever
  et~al\mbox{.}}{2014}]%
        {Seq2Seq}
\bibfield{author}{\bibinfo{person}{Ilya Sutskever}, \bibinfo{person}{Oriol
  Vinyals}, {and} \bibinfo{person}{Quoc~V. Le}.}
  \bibinfo{year}{2014}\natexlab{}.
\newblock \showarticletitle{Sequence to Sequence Learning with Neural
  Networks}. In \bibinfo{booktitle}{\emph{{NIPS}}}.
  \bibinfo{pages}{3104--3112}.
\newblock


\bibitem[\protect\citeauthoryear{Tedjopurnomo, Bao, Zheng, Choudhury, and
  Qin}{Tedjopurnomo et~al\mbox{.}}{2022}]%
        {trafficsurvey1}
\bibfield{author}{\bibinfo{person}{David~Alexander Tedjopurnomo},
  \bibinfo{person}{Zhifeng Bao}, \bibinfo{person}{Baihua Zheng},
  \bibinfo{person}{Farhana~Murtaza Choudhury}, {and} \bibinfo{person}{Alex~Kai
  Qin}.} \bibinfo{year}{2022}\natexlab{}.
\newblock \showarticletitle{A Survey on Modern Deep Neural Network for Traffic
  Prediction: Trends, Methods and Challenges}.
\newblock \bibinfo{journal}{\emph{{IEEE} Trans. Knowl. Data Eng.}}
  \bibinfo{volume}{34}, \bibinfo{number}{4} (\bibinfo{year}{2022}),
  \bibinfo{pages}{1544--1561}.
\newblock


\bibitem[\protect\citeauthoryear{van~den Oord, Dieleman, Zen, Simonyan,
  Vinyals, Graves, Kalchbrenner, Senior, and Kavukcuoglu}{van~den Oord
  et~al\mbox{.}}{2016}]%
        {WaveNet}
\bibfield{author}{\bibinfo{person}{A{\"{a}}ron van~den Oord},
  \bibinfo{person}{Sander Dieleman}, \bibinfo{person}{Heiga Zen},
  \bibinfo{person}{Karen Simonyan}, \bibinfo{person}{Oriol Vinyals},
  \bibinfo{person}{Alex Graves}, \bibinfo{person}{Nal Kalchbrenner},
  \bibinfo{person}{Andrew~W. Senior}, {and} \bibinfo{person}{Koray
  Kavukcuoglu}.} \bibinfo{year}{2016}\natexlab{}.
\newblock \showarticletitle{WaveNet: {A} Generative Model for Raw Audio}. In
  \bibinfo{booktitle}{\emph{{SSW}}}. \bibinfo{publisher}{{ISCA}},
  \bibinfo{pages}{125}.
\newblock


\bibitem[\protect\citeauthoryear{Wang, Lin, Guo, and Wan}{Wang
  et~al\mbox{.}}{2021b}]%
        {GSNet}
\bibfield{author}{\bibinfo{person}{Beibei Wang}, \bibinfo{person}{Youfang Lin},
  \bibinfo{person}{Shengnan Guo}, {and} \bibinfo{person}{Huaiyu Wan}.}
  \bibinfo{year}{2021}\natexlab{b}.
\newblock \showarticletitle{GSNet: Learning Spatial-Temporal Correlations from
  Geographical and Semantic Aspects for Traffic Accident Risk Forecasting}. In
  \bibinfo{booktitle}{\emph{{AAAI}}}. \bibinfo{publisher}{{AAAI} Press},
  \bibinfo{pages}{4402--4409}.
\newblock


\bibitem[\protect\citeauthoryear{Wang, Gu, Wu, Liu, and Xiong}{Wang
  et~al\mbox{.}}{2016}]%
        {ECRNN}
\bibfield{author}{\bibinfo{person}{Jingyuan Wang}, \bibinfo{person}{Qian Gu},
  \bibinfo{person}{Junjie Wu}, \bibinfo{person}{Guannan Liu}, {and}
  \bibinfo{person}{Zhang Xiong}.} \bibinfo{year}{2016}\natexlab{}.
\newblock \showarticletitle{Traffic Speed Prediction and Congestion Source
  Exploration: {A} Deep Learning Method}. In
  \bibinfo{booktitle}{\emph{{ICDM}}}. \bibinfo{publisher}{{IEEE} Computer
  Society}, \bibinfo{pages}{499--508}.
\newblock


\bibitem[\protect\citeauthoryear{Wang, Jiang, Jiang, Li, and Zhao}{Wang
  et~al\mbox{.}}{2021a}]%
        {libcity}
\bibfield{author}{\bibinfo{person}{Jingyuan Wang}, \bibinfo{person}{Jiawei
  Jiang}, \bibinfo{person}{Wenjun Jiang}, \bibinfo{person}{Chao Li}, {and}
  \bibinfo{person}{Wayne~Xin Zhao}.} \bibinfo{year}{2021}\natexlab{a}.
\newblock \showarticletitle{LibCity: An Open Library for Traffic Prediction}.
  In \bibinfo{booktitle}{\emph{{SIGSPATIAL/GIS}}}. \bibinfo{publisher}{{ACM}},
  \bibinfo{pages}{145--148}.
\newblock


\bibitem[\protect\citeauthoryear{Wang, Cao, and Yu}{Wang et~al\mbox{.}}{2022}]%
        {stsurvey}
\bibfield{author}{\bibinfo{person}{Senzhang Wang}, \bibinfo{person}{Jiannong
  Cao}, {and} \bibinfo{person}{Philip~S. Yu}.} \bibinfo{year}{2022}\natexlab{}.
\newblock \showarticletitle{Deep Learning for Spatio-Temporal Data Mining: {A}
  Survey}.
\newblock \bibinfo{journal}{\emph{{IEEE} Trans. Knowl. Data Eng.}}
  \bibinfo{volume}{34}, \bibinfo{number}{8} (\bibinfo{year}{2022}),
  \bibinfo{pages}{3681--3700}.
\newblock


\bibitem[\protect\citeauthoryear{Williams and Hoel}{Williams and Hoel}{2003}]%
        {ARIMA}
\bibfield{author}{\bibinfo{person}{Billy~M Williams} {and}
  \bibinfo{person}{Lester~A Hoel}.} \bibinfo{year}{2003}\natexlab{}.
\newblock \showarticletitle{Modeling and forecasting vehicular traffic flow as
  a seasonal ARIMA process: Theoretical basis and empirical results}.
\newblock \bibinfo{journal}{\emph{Journal of transportation engineering}}
  \bibinfo{volume}{129}, \bibinfo{number}{6} (\bibinfo{year}{2003}),
  \bibinfo{pages}{664--672}.
\newblock


\bibitem[\protect\citeauthoryear{Wu, Zhang, Guo, He, Yang, and Jensen}{Wu
  et~al\mbox{.}}{2021}]%
        {AutoCTS}
\bibfield{author}{\bibinfo{person}{Xinle Wu}, \bibinfo{person}{Dalin Zhang},
  \bibinfo{person}{Chenjuan Guo}, \bibinfo{person}{Chaoyang He},
  \bibinfo{person}{Bin Yang}, {and} \bibinfo{person}{Christian~S. Jensen}.}
  \bibinfo{year}{2021}\natexlab{}.
\newblock \showarticletitle{AutoCTS: Automated Correlated Time Series
  Forecasting}.
\newblock \bibinfo{journal}{\emph{Proc. {VLDB} Endow.}} \bibinfo{volume}{15},
  \bibinfo{number}{4} (\bibinfo{year}{2021}), \bibinfo{pages}{971--983}.
\newblock


\bibitem[\protect\citeauthoryear{Wu, Pan, Long, Jiang, Chang, and Zhang}{Wu
  et~al\mbox{.}}{2020}]%
        {MTGNN}
\bibfield{author}{\bibinfo{person}{Zonghan Wu}, \bibinfo{person}{Shirui Pan},
  \bibinfo{person}{Guodong Long}, \bibinfo{person}{Jing Jiang},
  \bibinfo{person}{Xiaojun Chang}, {and} \bibinfo{person}{Chengqi Zhang}.}
  \bibinfo{year}{2020}\natexlab{}.
\newblock \showarticletitle{Connecting the dots: Multivariate time series
  forecasting with graph neural networks}. In \bibinfo{booktitle}{\emph{KDD}}.
  \bibinfo{pages}{753--763}.
\newblock


\bibitem[\protect\citeauthoryear{Wu, Pan, Long, Jiang, and Zhang}{Wu
  et~al\mbox{.}}{2019}]%
        {GWNET}
\bibfield{author}{\bibinfo{person}{Z Wu}, \bibinfo{person}{S Pan},
  \bibinfo{person}{G Long}, \bibinfo{person}{J Jiang}, {and} \bibinfo{person}{C
  Zhang}.} \bibinfo{year}{2019}\natexlab{}.
\newblock \showarticletitle{Graph WaveNet for Deep Spatial-Temporal Graph
  Modeling}. In \bibinfo{booktitle}{\emph{IJCAI}}. International Joint
  Conferences on Artificial Intelligence Organization.
\newblock


\bibitem[\protect\citeauthoryear{Xu, Dai, Liu, Gao, Lin, Qi, and Xiong}{Xu
  et~al\mbox{.}}{2020}]%
        {STTN}
\bibfield{author}{\bibinfo{person}{Mingxing Xu}, \bibinfo{person}{Wenrui Dai},
  \bibinfo{person}{Chunmiao Liu}, \bibinfo{person}{Xing Gao},
  \bibinfo{person}{Weiyao Lin}, \bibinfo{person}{Guo{-}Jun Qi}, {and}
  \bibinfo{person}{Hongkai Xiong}.} \bibinfo{year}{2020}\natexlab{}.
\newblock \showarticletitle{Spatial-Temporal Transformer Networks for Traffic
  Flow Forecasting}.
\newblock \bibinfo{journal}{\emph{CoRR}}  \bibinfo{volume}{abs/2001.02908}
  (\bibinfo{year}{2020}).
\newblock


\bibitem[\protect\citeauthoryear{Yan, Ma, and Pu}{Yan et~al\mbox{.}}{2022}]%
        {TFormer}
\bibfield{author}{\bibinfo{person}{Haoyang Yan}, \bibinfo{person}{Xiaolei Ma},
  {and} \bibinfo{person}{Ziyuan Pu}.} \bibinfo{year}{2022}\natexlab{}.
\newblock \showarticletitle{Learning Dynamic and Hierarchical Traffic
  Spatiotemporal Features With Transformer}.
\newblock \bibinfo{journal}{\emph{{IEEE} Trans. Intell. Transp. Syst.}}
  \bibinfo{volume}{23}, \bibinfo{number}{11} (\bibinfo{year}{2022}),
  \bibinfo{pages}{22386--22399}.
\newblock


\bibitem[\protect\citeauthoryear{Yao, Tang, Wei, Zheng, and Li}{Yao
  et~al\mbox{.}}{2019}]%
        {STDN}
\bibfield{author}{\bibinfo{person}{Huaxiu Yao}, \bibinfo{person}{Xianfeng
  Tang}, \bibinfo{person}{Hua Wei}, \bibinfo{person}{Guanjie Zheng}, {and}
  \bibinfo{person}{Zhenhui Li}.} \bibinfo{year}{2019}\natexlab{}.
\newblock \showarticletitle{Revisiting Spatial-Temporal Similarity: {A} Deep
  Learning Framework for Traffic Prediction}. In
  \bibinfo{booktitle}{\emph{{AAAI}}}. \bibinfo{publisher}{{AAAI} Press},
  \bibinfo{pages}{5668--5675}.
\newblock


\bibitem[\protect\citeauthoryear{Yao, Wu, Ke, Tang, Jia, Lu, Gong, Ye, and
  Li}{Yao et~al\mbox{.}}{2018}]%
        {DMVSTNet}
\bibfield{author}{\bibinfo{person}{Huaxiu Yao}, \bibinfo{person}{Fei Wu},
  \bibinfo{person}{Jintao Ke}, \bibinfo{person}{Xianfeng Tang},
  \bibinfo{person}{Yitian Jia}, \bibinfo{person}{Siyu Lu},
  \bibinfo{person}{Pinghua Gong}, \bibinfo{person}{Jieping Ye}, {and}
  \bibinfo{person}{Zhenhui Li}.} \bibinfo{year}{2018}\natexlab{}.
\newblock \showarticletitle{Deep Multi-View Spatial-Temporal Network for Taxi
  Demand Prediction}. In \bibinfo{booktitle}{\emph{{AAAI}}}.
  \bibinfo{publisher}{{AAAI} Press}, \bibinfo{pages}{2588--2595}.
\newblock


\bibitem[\protect\citeauthoryear{Ye, Liu, Du, Sun, Li, Fu, and Xiong}{Ye
  et~al\mbox{.}}{2022b}]%
        {ESG}
\bibfield{author}{\bibinfo{person}{Junchen Ye}, \bibinfo{person}{Zihan Liu},
  \bibinfo{person}{Bowen Du}, \bibinfo{person}{Leilei Sun},
  \bibinfo{person}{Weimiao Li}, \bibinfo{person}{Yanjie Fu}, {and}
  \bibinfo{person}{Hui Xiong}.} \bibinfo{year}{2022}\natexlab{b}.
\newblock \showarticletitle{Learning the Evolutionary and Multi-scale Graph
  Structure for Multivariate Time Series Forecasting}. In
  \bibinfo{booktitle}{\emph{{KDD}}}. \bibinfo{publisher}{{ACM}},
  \bibinfo{pages}{2296--2306}.
\newblock


\bibitem[\protect\citeauthoryear{Ye, Sun, Du, Fu, and Xiong}{Ye
  et~al\mbox{.}}{2021}]%
        {CCRNN}
\bibfield{author}{\bibinfo{person}{Junchen Ye}, \bibinfo{person}{Leilei Sun},
  \bibinfo{person}{Bowen Du}, \bibinfo{person}{Yanjie Fu}, {and}
  \bibinfo{person}{Hui Xiong}.} \bibinfo{year}{2021}\natexlab{}.
\newblock \showarticletitle{Coupled Layer-wise Graph Convolution for
  Transportation Demand Prediction}. In \bibinfo{booktitle}{\emph{{AAAI}}}.
  \bibinfo{publisher}{{AAAI} Press}, \bibinfo{pages}{4617--4625}.
\newblock


\bibitem[\protect\citeauthoryear{Ye, Zhao, Ye, and Xu}{Ye
  et~al\mbox{.}}{2022c}]%
        {trafficsurvey2}
\bibfield{author}{\bibinfo{person}{Jiexia Ye}, \bibinfo{person}{Juanjuan Zhao},
  \bibinfo{person}{Kejiang Ye}, {and} \bibinfo{person}{Cheng{-}Zhong Xu}.}
  \bibinfo{year}{2022}\natexlab{c}.
\newblock \showarticletitle{How to Build a Graph-Based Deep Learning
  Architecture in Traffic Domain: {A} Survey}.
\newblock \bibinfo{journal}{\emph{{IEEE} Trans. Intell. Transp. Syst.}}
  \bibinfo{volume}{23}, \bibinfo{number}{5} (\bibinfo{year}{2022}),
  \bibinfo{pages}{3904--3924}.
\newblock


\bibitem[\protect\citeauthoryear{Ye, Fang, Sun, Zhang, and Xiang}{Ye
  et~al\mbox{.}}{2022a}]%
        {MGT}
\bibfield{author}{\bibinfo{person}{Xue Ye}, \bibinfo{person}{Shen Fang},
  \bibinfo{person}{Fang Sun}, \bibinfo{person}{Chunxia Zhang}, {and}
  \bibinfo{person}{Shiming Xiang}.} \bibinfo{year}{2022}\natexlab{a}.
\newblock \showarticletitle{Meta Graph Transformer: {A} Novel Framework for
  Spatial-Temporal Traffic Prediction}.
\newblock \bibinfo{journal}{\emph{Neurocomputing}}  \bibinfo{volume}{491}
  (\bibinfo{year}{2022}), \bibinfo{pages}{544--563}.
\newblock


\bibitem[\protect\citeauthoryear{Yin, Wu, Wei, Shen, Qi, and Yin}{Yin
  et~al\mbox{.}}{2021}]%
        {yin2021deep}
\bibfield{author}{\bibinfo{person}{Xueyan Yin}, \bibinfo{person}{Genze Wu},
  \bibinfo{person}{Jinze Wei}, \bibinfo{person}{Yanming Shen},
  \bibinfo{person}{Heng Qi}, {and} \bibinfo{person}{Baocai Yin}.}
  \bibinfo{year}{2021}\natexlab{}.
\newblock \showarticletitle{Deep Learning on Traffic Prediction: Methods,
  Analysis and Future Directions}.
\newblock \bibinfo{journal}{\emph{IEEE Transactions on Intelligent
  Transportation Systems}} (\bibinfo{year}{2021}).
\newblock


\bibitem[\protect\citeauthoryear{Yu, Yin, and Zhu}{Yu et~al\mbox{.}}{2018}]%
        {STGCN}
\bibfield{author}{\bibinfo{person}{Bing Yu}, \bibinfo{person}{Haoteng Yin},
  {and} \bibinfo{person}{Zhanxing Zhu}.} \bibinfo{year}{2018}\natexlab{}.
\newblock \showarticletitle{Spatio-temporal Graph Convolutional Networks: A
  Deep Learning Framework for Traffic Forecasting}. In
  \bibinfo{booktitle}{\emph{Proceedings of the 27th International Joint
  Conference on Artificial Intelligence (IJCAI)}}.
\newblock


\bibitem[\protect\citeauthoryear{Yuan, Zheng, Xie, and Sun}{Yuan
  et~al\mbox{.}}{2011}]%
        {yuan2011driving}
\bibfield{author}{\bibinfo{person}{Jing Yuan}, \bibinfo{person}{Yu Zheng},
  \bibinfo{person}{Xing Xie}, {and} \bibinfo{person}{Guangzhong Sun}.}
  \bibinfo{year}{2011}\natexlab{}.
\newblock \showarticletitle{Driving with knowledge from the physical world}. In
  \bibinfo{booktitle}{\emph{Proceedings of the 17th ACM SIGKDD international
  conference on Knowledge discovery and data mining}}.
  \bibinfo{pages}{316--324}.
\newblock


\bibitem[\protect\citeauthoryear{Yuan, Zheng, Zhang, Xie, Xie, Sun, and
  Huang}{Yuan et~al\mbox{.}}{2010}]%
        {yuan2010t}
\bibfield{author}{\bibinfo{person}{Jing Yuan}, \bibinfo{person}{Yu Zheng},
  \bibinfo{person}{Chengyang Zhang}, \bibinfo{person}{Wenlei Xie},
  \bibinfo{person}{Xing Xie}, \bibinfo{person}{Guangzhong Sun}, {and}
  \bibinfo{person}{Yan Huang}.} \bibinfo{year}{2010}\natexlab{}.
\newblock \showarticletitle{T-drive: driving directions based on taxi
  trajectories}. In \bibinfo{booktitle}{\emph{Proceedings of the 18th
  SIGSPATIAL International conference on advances in geographic information
  systems}}. \bibinfo{pages}{99--108}.
\newblock


\bibitem[\protect\citeauthoryear{Zang, Han, Li, Wan, and Wang}{Zang
  et~al\mbox{.}}{2022}]%
        {POIrec}
\bibfield{author}{\bibinfo{person}{Hongyu Zang}, \bibinfo{person}{Dongcheng
  Han}, \bibinfo{person}{Xin Li}, \bibinfo{person}{Zhifeng Wan}, {and}
  \bibinfo{person}{Mingzhong Wang}.} \bibinfo{year}{2022}\natexlab{}.
\newblock \showarticletitle{{CHA:} Categorical Hierarchy-based Attention for
  Next {POI} Recommendation}.
\newblock \bibinfo{journal}{\emph{{ACM} Trans. Inf. Syst.}}
  \bibinfo{volume}{40}, \bibinfo{number}{1} (\bibinfo{year}{2022}),
  \bibinfo{pages}{7:1--7:22}.
\newblock


\bibitem[\protect\citeauthoryear{Zhang, Chen, Cui, Guo, and Zhu}{Zhang
  et~al\mbox{.}}{2020b}]%
        {ResLSTM}
\bibfield{author}{\bibinfo{person}{Jinlei Zhang}, \bibinfo{person}{Feng Chen},
  \bibinfo{person}{Zhiyong Cui}, \bibinfo{person}{Yinan Guo}, {and}
  \bibinfo{person}{Yadi Zhu}.} \bibinfo{year}{2020}\natexlab{b}.
\newblock \showarticletitle{Deep learning architecture for short-term passenger
  flow forecasting in urban rail transit}.
\newblock \bibinfo{journal}{\emph{IEEE Transactions on Intelligent
  Transportation Systems}} (\bibinfo{year}{2020}).
\newblock


\bibitem[\protect\citeauthoryear{Zhang, Shi, Xie, Ma, King, and Yeung}{Zhang
  et~al\mbox{.}}{2018}]%
        {GaAN}
\bibfield{author}{\bibinfo{person}{Jiani Zhang}, \bibinfo{person}{Xingjian
  Shi}, \bibinfo{person}{Junyuan Xie}, \bibinfo{person}{Hao Ma},
  \bibinfo{person}{Irwin King}, {and} \bibinfo{person}{Dit{-}Yan Yeung}.}
  \bibinfo{year}{2018}\natexlab{}.
\newblock \showarticletitle{GaAN: Gated Attention Networks for Learning on
  Large and Spatiotemporal Graphs}. In \bibinfo{booktitle}{\emph{{UAI}}}.
  \bibinfo{publisher}{{AUAI} Press}, \bibinfo{pages}{339--349}.
\newblock


\bibitem[\protect\citeauthoryear{Zhang, Zheng, and Qi}{Zhang
  et~al\mbox{.}}{2017}]%
        {STResNet}
\bibfield{author}{\bibinfo{person}{Junbo Zhang}, \bibinfo{person}{Yu Zheng},
  {and} \bibinfo{person}{Dekang Qi}.} \bibinfo{year}{2017}\natexlab{}.
\newblock \showarticletitle{Deep spatio-temporal residual networks for citywide
  crowd flows prediction}. In \bibinfo{booktitle}{\emph{Proceedings of the AAAI
  Conference on Artificial Intelligence}}, Vol.~\bibinfo{volume}{31}.
\newblock


\bibitem[\protect\citeauthoryear{Zhang, Chang, Meng, Xiang, and Pan}{Zhang
  et~al\mbox{.}}{2020a}]%
        {SLCNN}
\bibfield{author}{\bibinfo{person}{Qi Zhang}, \bibinfo{person}{Jianlong Chang},
  \bibinfo{person}{Gaofeng Meng}, \bibinfo{person}{Shiming Xiang}, {and}
  \bibinfo{person}{Chunhong Pan}.} \bibinfo{year}{2020}\natexlab{a}.
\newblock \showarticletitle{Spatio-Temporal Graph Structure Learning for
  Traffic Forecasting}. In \bibinfo{booktitle}{\emph{{AAAI}}}.
  \bibinfo{publisher}{{AAAI} Press}, \bibinfo{pages}{1177--1185}.
\newblock


\bibitem[\protect\citeauthoryear{Zhang, Huang, Xu, Xia, Dai, Bo, Zhang, and
  Zheng}{Zhang et~al\mbox{.}}{2021}]%
        {STGDN}
\bibfield{author}{\bibinfo{person}{Xiyue Zhang}, \bibinfo{person}{Chao Huang},
  \bibinfo{person}{Yong Xu}, \bibinfo{person}{Lianghao Xia},
  \bibinfo{person}{Peng Dai}, \bibinfo{person}{Liefeng Bo},
  \bibinfo{person}{Junbo Zhang}, {and} \bibinfo{person}{Yu Zheng}.}
  \bibinfo{year}{2021}\natexlab{}.
\newblock \showarticletitle{Traffic Flow Forecasting with Spatial-Temporal
  Graph Diffusion Network}. In \bibinfo{booktitle}{\emph{{AAAI}}}.
  \bibinfo{publisher}{{AAAI} Press}, \bibinfo{pages}{15008--15015}.
\newblock


\bibitem[\protect\citeauthoryear{Zhao, Song, Zhang, Liu, Wang, Lin, Deng, and
  Li}{Zhao et~al\mbox{.}}{2019}]%
        {TGCN}
\bibfield{author}{\bibinfo{person}{Ling Zhao}, \bibinfo{person}{Yujiao Song},
  \bibinfo{person}{Chao Zhang}, \bibinfo{person}{Yu Liu}, \bibinfo{person}{Pu
  Wang}, \bibinfo{person}{Tao Lin}, \bibinfo{person}{Min Deng}, {and}
  \bibinfo{person}{Haifeng Li}.} \bibinfo{year}{2019}\natexlab{}.
\newblock \showarticletitle{T-gcn: A temporal graph convolutional network for
  traffic prediction}.
\newblock \bibinfo{journal}{\emph{IEEE Transactions on Intelligent
  Transportation Systems}} \bibinfo{volume}{21}, \bibinfo{number}{9}
  (\bibinfo{year}{2019}), \bibinfo{pages}{3848--3858}.
\newblock


\bibitem[\protect\citeauthoryear{Zheng, Fan, Wang, and Qi}{Zheng
  et~al\mbox{.}}{2020}]%
        {GMAN}
\bibfield{author}{\bibinfo{person}{Chuanpan Zheng}, \bibinfo{person}{Xiaoliang
  Fan}, \bibinfo{person}{Cheng Wang}, {and} \bibinfo{person}{Jianzhong Qi}.}
  \bibinfo{year}{2020}\natexlab{}.
\newblock \showarticletitle{{GMAN:} {A} Graph Multi-Attention Network for
  Traffic Prediction}. In \bibinfo{booktitle}{\emph{{AAAI}}}.
  \bibinfo{publisher}{{AAAI} Press}, \bibinfo{pages}{1234--1241}.
\newblock


\bibitem[\protect\citeauthoryear{Zhou, Jin, Zhang, Qin, Jiao, Wang, Wu, Yu, and
  Ye}{Zhou et~al\mbox{.}}{2019}]%
        {dispatching}
\bibfield{author}{\bibinfo{person}{Ming Zhou}, \bibinfo{person}{Jiarui Jin},
  \bibinfo{person}{Weinan Zhang}, \bibinfo{person}{Zhiwei~(Tony) Qin},
  \bibinfo{person}{Yan Jiao}, \bibinfo{person}{Chenxi Wang},
  \bibinfo{person}{Guobin Wu}, \bibinfo{person}{Yong Yu}, {and}
  \bibinfo{person}{Jieping Ye}.} \bibinfo{year}{2019}\natexlab{}.
\newblock \showarticletitle{Multi-Agent Reinforcement Learning for
  Order-dispatching via Order-Vehicle Distribution Matching}. In
  \bibinfo{booktitle}{\emph{{CIKM}}}. \bibinfo{publisher}{{ACM}},
  \bibinfo{pages}{2645--2653}.
\newblock

\bibitem[\protect\citeauthoryear{Wang, Cui, Wang, Pei, Zhu, and Yang}{Wang
  et~al\mbox{.}}{2017}]%
        {wang2017community}
\bibfield{author}{\bibinfo{person}{Xiao Wang}, \bibinfo{person}{Peng Cui},
  \bibinfo{person}{Jing Wang}, \bibinfo{person}{Jian Pei},
  \bibinfo{person}{Wenwu Zhu}, {and} \bibinfo{person}{Shiqiang Yang}.}
  \bibinfo{year}{2017}\natexlab{}.
\newblock \showarticletitle{Community preserving network embedding}. In
  \bibinfo{booktitle}{\emph{Proceedings of the AAAI conference on artificial
  intelligence}}, Vol.~\bibinfo{volume}{31}.
\newblock

\end{thebibliography}

\end{document}